\documentclass[runningheads]{llncs}

\usepackage{accv}

\usepackage{accvabbrv}

\usepackage[table]{xcolor}

\usepackage{graphicx}
\usepackage{booktabs}

\usepackage{multirow}
\usepackage{placeins}

\usepackage[accsupp]{axessibility}  %

\usepackage{hyperref}

\usepackage{orcidlink}

\begin{document}

\title{MaRO-GS: Mask-Robust Object-Centric Gaussian Splatting from Inconsistent Multi-view Masks}

\titlerunning{MaRO-GS}

\author{Eunji Kim$^{*}$\and
Gahyeon Kim$^{*}$ \and
Gianella Cravioto \and
Dong-hun Lee \and
Chaewon~Moon \and
Chae-yeong Song \and
Sang-hyo Park$^{\dagger}$
}

\authorrunning{E.~Kim et al.}

\institute{School of Computer Science and Engineering\\
Kyungpook National University, Daegu, Republic of Korea\\
\email{\{now0104, hwang190190, gianellacrv9, hy05205,\\
moonchaewon2, yeongsong, s.park\}@knu.ac.kr}\\
\url{https://eunjikim02.github.io/marogs/}
}

\maketitle

\let\thefootnote\relax
\footnotetext{$^{*}$Equal contribution\\
$^{\dagger}$Corresponding author}

\begin{figure}[t]
    \centering
    \includegraphics[width=0.9\linewidth]{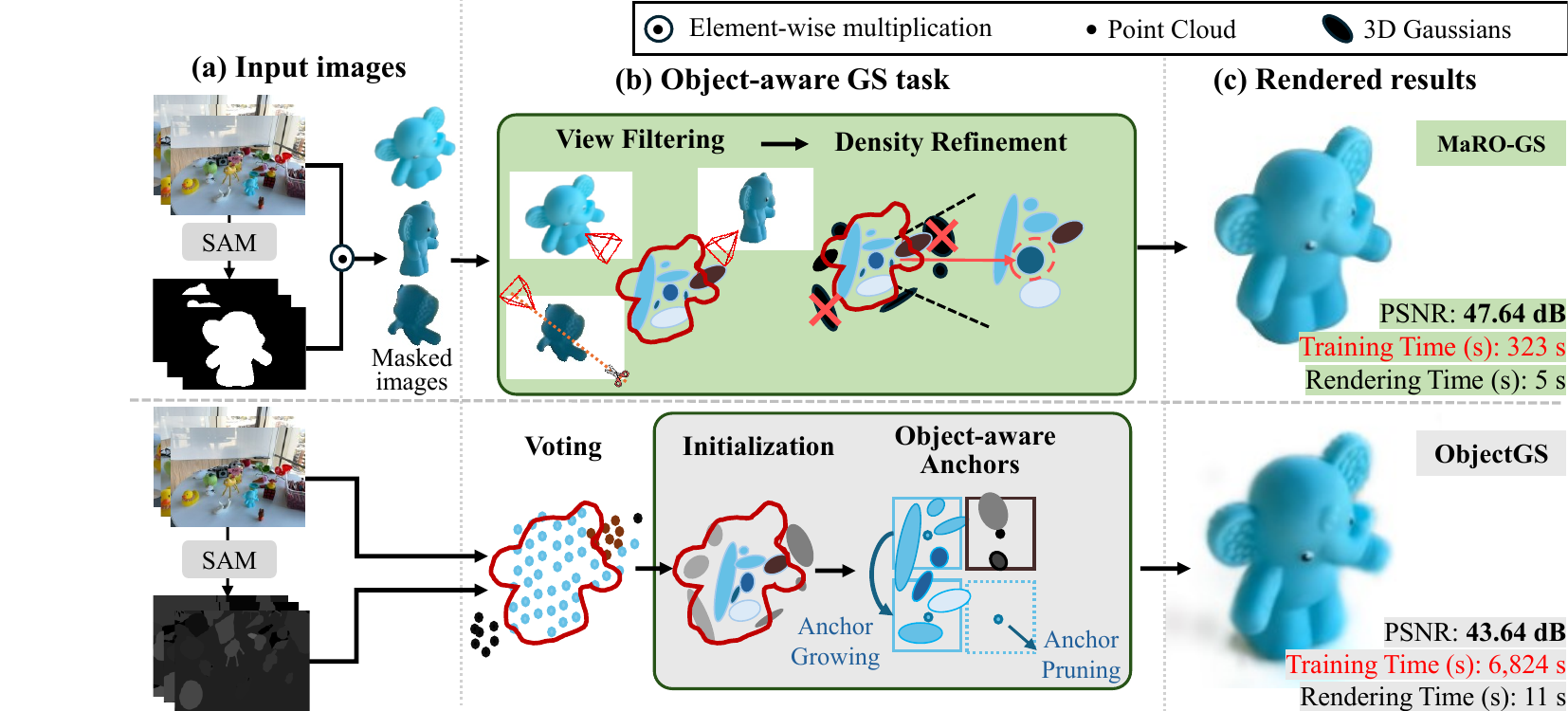}
    \caption{Comparison between MaRO-GS and ObjectGS on a single object from LERF-Mask. (a) MaRO-GS directly reconstructs the target object from object-masked multi-view images (top), whereas ObjectGS reconstructs the entire scene with its native full-scene pipeline before extracting the target object (bottom). This leads to longer training time (6,824 s vs.\ 323 s).}
    \label{fig:fullscene}
\end{figure}

\begin{abstract}
We address the challenge of accurate 3D object reconstruction from multi-view images in Gaussian Splatting. Existing object-level 3DGS methods reconstruct the entire scene rather than directly optimizing the target object, even when only the target object is needed, which incurs substantial computational overhead. They also rely on 2D segmentation masks to associate Gaussians with objects, but these masks are often inconsistent across views. Such inconsistencies corrupt Gaussian optimization and produce incorrectly supervised Gaussians that degrade object reconstruction fidelity. To overcome these limitations, we propose MaRO-GS, a 3DGS framework that directly optimizes target-object Gaussians from object-masked multi-view images and remains robust to inconsistent supervision. For reliable supervision, mask-reliability view filtering excludes unreliable views. Object-supported Gaussian density control suppresses Gaussians irrelevant to the target object and prevents background densification, while Silhouette-aligned Object Loss maintains object-focused optimization. Extensive experiments across diverse datasets demonstrate that MaRO-GS improves PSNR, segmentation accuracy, and computational efficiency, with the largest PSNR gain of 2.05 dB on the small-object LERF-Mask dataset.
\end{abstract}

\section{Introduction}
\label{sec:intro}
3D object reconstruction is essential for applications such as object editing~\cite{xie2026objectmorpher,yenphraphai2024image,ni2025decompositional}, AR/VR~\cite{shen2025gaussianshopvr}, and 3D content creation~\cite{ni2025decompositional}. Despite the importance of 3D object reconstruction, recent 3D Gaussian Splatting (3DGS)~\cite{3dgs} research has primarily focused on improving scene-level reconstruction quality~\cite{zeng2025frequency,chao2025textured,zhang2024fregs} and computational efficiency~\cite{pup, hanson2025speedysplat,dai2025efficient,radsplat}. To support object-centric reconstruction, several methods have extended 3DGS to enable object-level reconstruction~\cite{Ye2024GaussianGrouping, Zhu2025ObjectGS, zhang2025cob, mazzucchelli2026bea, hu2026sagd, jain2024gaussiancut}. However, as illustrated in ~\cref{fig:fullscene}, these approaches reconstruct the entire scene before extracting the target object, which incurs unnecessary computational overhead. Furthermore, they rely on SAM\cite{kirillov2023segment}-generated masks that are inconsistent across views, thereby destabilizing Gaussian optimization and degrading object reconstruction quality.

To improve object extraction quality, recent methods propose boundary refinement strategies~\cite{zhang2025cob, mazzucchelli2026bea, hu2026sagd} or enforce cross-view mask consistency via tracking and voting~\cite{Ye2024GaussianGrouping, Zhu2025ObjectGS}. 
The former refines object boundaries to mitigate extraction artifacts, yet reconstructing the entire scene still incurs substantial computational overhead.
The latter ensures cross-view object identity consistency but does not assess individual mask reliability, so inaccurate masks are still used as supervision during optimization.
Meanwhile, Trace3D~\cite{shen2025trace3d} identifies inconsistent masks via instance weight tracing but corrects rather than filters inaccurate masks, which allows erroneous supervision into the optimization process.

To address these challenges, we propose \textbf{MaRO-GS} (\textbf{Ma}sk-\textbf{R}obust \textbf{O}bject-centric \textbf{G}aussian \textbf{S}platting), a framework for direct optimization of target-object Gaussians from object-masked multi-view images under inconsistent mask supervision. 
Mask-reliability view filtering establishes reliable mask supervision by identifying and excluding unreliable views during training.
 Inconsistent mask supervision can also lead to unnecessary Gaussian densification around object boundaries and background regions. Object-supported Gaussian density control suppresses densification outside the target object and removes Gaussians irrelevant to the object to maintain a compact object-centric representation while improving reconstruction quality.
In addition, the silhouette-aligned object loss further constrains optimization with object-focused photometric and silhouette supervision, encouraging the reconstructed object to align with the target object silhouette.

Extensive experiments demonstrate that MaRO-GS improves PSNR over existing methods, with the largest gain of 2.05 dB on the small-object LERF-Mask dataset~\cite{Ye2024GaussianGrouping} while directly reconstructing the target object without scene-level reconstruction. As a complementary analysis, we evaluate target object reconstruction quality from object-centric viewpoints.

\begin{itemize}
    \item We propose MaRO-GS, a mask-robust framework for object-centric 3DGS that directly optimizes target-object Gaussians from object-masked multi-view observations under inconsistent mask supervision.
     \item We propose mask-reliability view filtering, which stabilizes object-centric optimization by identifying and excluding unreliable views during training.
    \item We introduce object-supported Gaussian density control, which removes Gaussians irrelevant to the target object and constrains densification for a compact, high-fidelity object representation.
    \item We provide object-centric and mask-error robustness evaluations that complement conventional scene-level evaluation for assessing target-object reconstruction quality, particularly under small object regions and inconsistent masks.
\end{itemize}

\section{Related Work}
\label{sec:relatedwork}
\subsection{Scene-level Reconstruction in 3DGS}
Recent 3DGS~\cite{3dgs} research has substantially improved scene-level reconstruction fidelity through enriched Gaussian representations and adaptive density control~\cite{lu2024scaffold,kheradmand20243d,ye2024absgs,chao2025textured,zeng2025frequency}. Other studies have instead focused on improving computational efficiency through pruning and compression techniques~\cite{LightGaussian,minisplatting,pup,eagles,C3DGS,hanson2025speedysplat,niedermayr2024compressed,radsplat}. However, these scene-level 3DGS methods are designed for entire scene reconstruction and do not account for individual objects within the scene.

\subsection{Object-level Reconstruction in 3DGS}
Several methods~\cite{Ye2024GaussianGrouping, Zhu2025ObjectGS, zhang2025cob, shen2025trace3d, mazzucchelli2026bea} have extended 3DGS to support object-level reconstruction. For example, ObjectGS\cite{Zhu2025ObjectGS} introduces object-aware anchors, COB-GS\cite{zhang2025cob} and Trace3D\cite{shen2025trace3d} address boundary ambiguity through Gaussian refinement, and BEA-GS\cite{mazzucchelli2026bea} improves object extraction via boundary and occupancy losses. 
However, these methods perform scene-level Gaussian optimization prior to object extraction, which results in unnecessary computational overhead. Since scene-level reconstruction optimizes Gaussian primitives for the entire scene, Gaussians near object boundaries receive supervision from both object and background regions. As a result, ambiguous Gaussians emerge, which limits object extraction quality despite various boundary refinement strategies~\cite{zhang2025cob, shen2025trace3d, mazzucchelli2026bea}.
Unlike these approaches, MaRO-GS  directly reconstructs the target object from object-masked multi-view images and applies object-supported Gaussian density control to better constrain Gaussian optimization around the target object.

\subsection{2D Mask Supervision in 3DGS}
Recent object-aware methods have leveraged 2D masks predicted by SAM\cite{kirillov2023segment} and lifted them into 3D representations~\cite{Ye2024GaussianGrouping,jang2025identity, shen2024flashsplat, yang2024gaussianobject,hu2026sagd, jain2024gaussiancut}. This enables applications such as object extraction, segmentation, and scene editing in 3DGS.
Although SAM provides powerful segmentation capabilities, it produces inconsistent masks across views for the same object. These inconsistencies manifest in both object boundaries\cite{zhang2025cob, shen2025trace3d, mazzucchelli2026bea} and mask granularity\cite{choi2024click}. They introduce conflicting supervision signals into the reconstruction process. As a result, object-centric Gaussian optimization becomes unreliable and reconstruction quality degrades.
To address this, Gaussian Grouping~\cite{Ye2024GaussianGrouping} and ObjectGS~\cite{Zhu2025ObjectGS} adopt different strategies to establish cross-view mask consistency. Specifically, Gaussian Grouping employs the DEVA\cite{cheng2023tracking} tracker for cross-view object ID tracking, while ObjectGS uses majority voting to assign object IDs. In both cases, the focus remains on cross-view object ID consistency rather than assessing the reliability of individual 2D mask predictions. Consequently, inaccurate masks can still be used as supervision during optimization.
Trace3D\cite{shen2025trace3d} also addresses multi-view mask inconsistency via GIT-based instance weight tracing. However, as it is designed for multi-object segmentation within a fully reconstructed scene, its consistency enforcement operates across all objects simultaneously rather than explicitly identifying and filtering low-quality masks for a target object.
Despite their differences, these methods do not explicitly filter inconsistent masks to ensure clean supervision for object-centric reconstruction. In contrast, MaRO-GS  removes unreliable views through mask-reliability view filtering and provides cleaner supervision.

\begin{figure*}[t]
    \centering
    \includegraphics[width=1.0\linewidth]{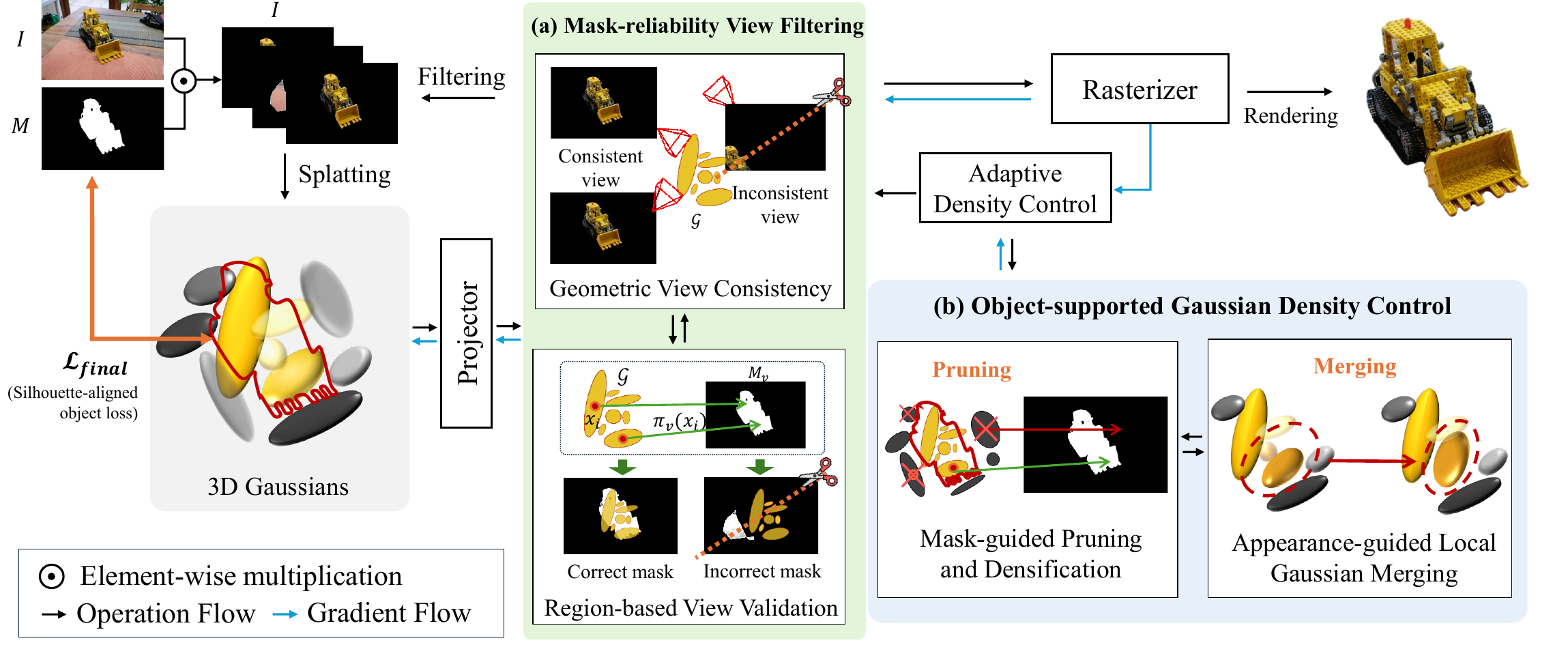}
    \caption{Overview of MaRO-GS  for object-centric 3D Gaussian Splatting. The pipeline first removes inconsistent views through (a) mask-reliability view filtering. It then suppresses Gaussians irrelevant to the target object via (b) object-supported Gaussian density control to improve object reconstruction fidelity.}
    \label{fig:framework}
\end{figure*}

\section{Methods}
We propose \textbf{MaRO-GS} (\textbf{Ma}sk-\textbf{R}obust \textbf{O}bject-centric \textbf{G}aussian \textbf{S}platting), illustrated in \cref{fig:framework}. 
Our method first formalizes the object-centric reconstruction task (\cref{sec:task}). To ensure reliable optimization under imperfect mask supervision, we introduce \textit{mask-reliability view filtering}, which enforces geometric consistency across training views (\cref{sec:filtering}). We further propose \textit{object-supported Gaussian density control} to suppress background Gaussians and concentrate density around target objects (\cref{sec:pruning}). Finally, we design a \textit{Silhouette-aligned object loss function} that prioritizes object regions during optimization (\cref{sec:oc_loss}). This unified design transforms conventional scene-level 3DGS into an object-centric reconstruction pipeline for robust and high-fidelity results under inconsistent mask supervision.

\subsection{Object-centric 3D Gaussian Splatting}
\label{sec:task}

We formalize object-centric 3DGS for accurate and efficient reconstruction of a target object by directly performing Gaussian optimization on object-masked observations, without reconstructing the entire scene. 
Specifically, given an RGB image $I$ and binary object mask $M$, we construct an object-masked observation:
\begin{equation}
    \tilde{I} = I \odot M,
    \label{eq:masked_input}
\end{equation}
where $\odot$ denotes element-wise multiplication.
This masking operation removes irrelevant scene regions and focuses the optimization on the target object.
However, simply applying object masks is insufficient, as optimization becomes directly dependent on their quality and consistency. When masks are inaccurate or inconsistent across views, the same 3D region may be supervised as foreground in one view and background in another, which corrupts Gaussian optimization and adaptive density control. Therefore, object-centric 3DGS requires robust optimization under imperfect mask supervision, which motivates the designs in the following sections.

\subsection{Mask-Reliability View Filtering}
\label{sec:filtering}

\begin{figure}[!t]
    \centering
    \includegraphics[width=0.7\linewidth]{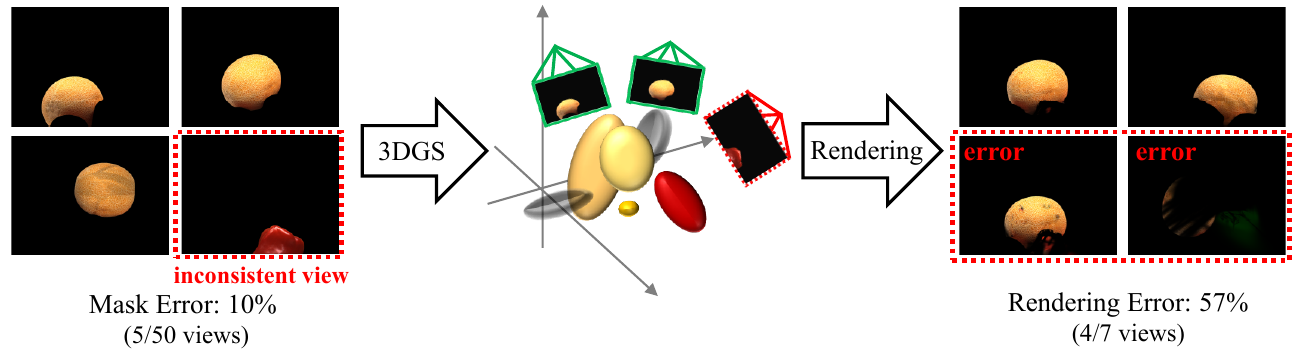}
    \caption{Illustration of the instability of conventional 3DGS under inconsistent masks.}
    \label{fig:problem}
\end{figure}

Since MaRO-GS  directly optimizes an object-centric Gaussian representation from masked observations, the reliability of each object mask directly determines the quality of supervision. 
However, segmentation models such as SAM~\cite{kirillov2023segment} often produce view-dependent mask errors, including missing object regions, boundary inaccuracies, and background inclusions. 
As shown in \cref{fig:problem}, even a small number of erroneous masks can disturb Gaussian optimization and lead to rendering artifacts in multiple views. 
This indicates that mask errors are not isolated 2D supervision noise, but can propagate through the shared 3D Gaussian representation. 
We therefore propose \textit{mask-reliability view filtering}, which identifies unreliable views and excludes them from subsequent training. 
This filtering is realized through two complementary stages: \textit{Geometric view consistency} and \textit{Region-based view validation}. 

\subsubsection{Geometric View Consistency.}
\label{sec:geometricview}
The first stage identifies geometrically inconsistent views that provide unreliable object-level supervision. Unlike scene-level view consistency, which treats all visible Gaussians equally, our goal is to evaluate whether the visible Gaussians are consistently supported by the target-object masks. 
For each view $v_i$, we render the current Gaussian set $\mathcal{G}$ and obtain its visible Gaussian set $\mathcal{V}_i$. 
For a Gaussian $g_k$, we define its object-supported visibility weight in view $v_i$ as
\begin{equation}
    \mathcal{V}_{\mathrm{obj}}^{k}(i)
    =
    \mathbf{1}\!\left[g_k \in \mathcal{V}_i\right]
    \cdot
    M_i\!\left(\pi_i(\mathbf{x}_k)\right)
    \cdot
    \alpha_k ,
    \label{eq:object_visibility}
\end{equation}

\noindent
where $\mathbf{x}_k$ and $\alpha_k$ are the 3D center and opacity of Gaussian $g_k$, respectively. $\pi_i(\mathbf{x}_k)$ denotes its projected 2D location in view $v_i$, and $M_i(\cdot)$ is the binary object mask for view $v_i$.
Building on the covisibility formulation of LongSplat~\cite{lin2025longsplat}, we introduce an object-supported covisibility measure.
Let $\mathcal{N}_i$ denote the neighboring view set of $v_i$, selected according to camera-center distance. 
The object covisibility score $\mathrm{C}_{i}$ is defined as
\begin{equation}
    \mathrm{C}_i
    =
    \frac{1}{K}
    \sum_{j \in \operatorname{TopK}(\mathcal{N}_i)}
    \frac{
    \sum_k
    \min\!\left(
    \mathcal{V}_{\mathrm{obj}}^{k}(i),
    \mathcal{V}_{\mathrm{obj}}^{k}(j)
    \right)
    }{
    \sum_k
    \max\!\left(
    \mathcal{V}_{\mathrm{obj}}^{k}(i),
    \mathcal{V}_{\mathrm{obj}}^{k}(j)
    \right)
    + \epsilon
    },
    \label{eq:covscore}
\end{equation}
where $\mathcal{N}_i$ denotes the pose-nearest candidate view set selected by camera-center distance, and $\operatorname{TopK}(\mathcal{N}_i)$ selects the three views in $\mathcal{N}_i$ that produce the highest object-supported covisibility responses with respect to the reference view $v_i$.
After computing $\mathrm{C}_i$ for all training views, we sort the scores as 
$\mathrm{C}_{(1)} \leq \cdots \leq \mathrm{C}_{(N)}$. 
We then determine $\tau_{\mathrm{geo}}$ from the largest score gap:
\begin{equation}
    s^* = \arg\max_s \left(\mathrm{C}_{(s+1)}-\mathrm{C}_{(s)}\right),  \quad
    \tau_{\mathrm{geo}} = \frac{\mathrm{C}_{(s^*)}+\mathrm{C}_{(s^*+1)}}{2}.
    \label{eq:adaptive_threshold}
\end{equation}
Rather than using a fixed absolute threshold, $\tau_{\mathrm{geo}}$ is determined from the relative score distribution, and the split is accepted only when the low-score cluster is sufficiently small and clearly separated from the remaining views.
The geometrically unreliable view set is then defined as
\begin{equation}
    \mathcal{V}_{\mathrm{bad}}^{\mathrm{geo}}
    =
    \left\{
    v_i
    \mid
    \mathrm{C}_i < \tau_{\mathrm{geo}}
    \right\}.
    \label{eq:bad_geo_views}
\end{equation}
These views are excluded from subsequent training, preventing inconsistent object-level supervision from contaminating the densification process.

\subsubsection{Region-based View Validation.} 
\label{sec:region}
Geometric consistency alone cannot catch views where the covisibility is acceptable but the 2D mask is spatially misaligned due to occlusion or segmentation failure. 
For these cases, we evaluate whether the current Gaussian distribution is well supported by the object mask in each view. 
Specifically, we compute a per-view hit ratio:

\begin{equation}
    H_i =
    \frac{1}{|\mathcal{G}_i|}
    \sum_{\mathbf{x}_k \in \mathcal{G}_i}
    M_i\!\bigl(\pi_i(\mathbf{x}_k)\bigr),
    \label{eq:hit_ratio}
\end{equation}
where $\mathcal{G}_i$ denotes the set of Gaussian centers whose projections lie inside the image domain of view $v_i$. 
Intuitively, $H_i$ measures the fraction of valid Gaussian projections that fall inside the object mask. 
Views with $H_i < \tau_{\mathrm{hit}}$ are added to $\mathcal{V}_{\mathrm{bad}}^{\mathrm{reg}}$ and excluded from subsequent training.
The final unreliable view set is defined as 
$\mathcal{V}_{\mathrm{bad}}=\mathcal{V}_{\mathrm{bad}}^{\mathrm{geo}}\cup\mathcal{V}_{\mathrm{bad}}^{\mathrm{reg}}$.

\subsection{Object-supported Gaussian Density Control} \label{sec:pruning}

While mask-reliability view filtering improves the reliability of view-level supervision, object-centric reconstruction also requires density-level control over the Gaussian representation. 
The Gaussian distribution should be allocated and maintained in regions that receive consistent target-object support across views. 
Without this constraint, Gaussians near object boundaries or residual background regions can persist during optimization and may be further amplified by adaptive densification. 
These unsupported Gaussians degrade object compactness and introduce residual background artifacts. 
To address this issue, we propose \textit{object-supported Gaussian density control}, which refines the Gaussian distribution through two complementary operations: \textit{Mask-guided Pruning and Densification} and \textit{Appearance-guided Local Gaussian Merging}.

\subsubsection{Mask-Guided Pruning and Densification.}

To remove such unsupported primitives, we measure how consistently each Gaussian is supported by object masks across views. 
For a Gaussian \(g_i\) with center \(\mathbf{x}_i\), let \(\Omega_i\) denote the set of views in which \(g_i\) is projected inside the image domain. We define its mask-overlap ratio as
\begin{equation}
\mathcal{P}_i =
\frac{1}{|\Omega_i|}
\sum_{v \in \Omega_i}
M_v\!\left(\pi_v(\mathbf{x}_i)\right).
\label{eq:mask_overlap}
\end{equation}
A low $\mathcal{P}_i$ indicates weak object-mask support and suggests that the Gaussian corresponds to background or boundary artifacts.
We adaptively determine the pruning threshold from the distribution of $\{\mathcal{P}_i\}$. 
Let $\mu$ and $\sigma$ denote the mean and standard deviation of mask-overlap ratios.
The threshold $\tau_{\mathrm{mask}}$ is derived from these statistics, with $\mu+k\sigma$ as a base criterion and a conservative upper bound to prevent over-pruning. 
We then remove unsupported Gaussians as
\begin{equation}
    \mathcal{G}_{\mathrm{prune}}
    =
    \{g_i \in \mathcal{G} \mid \mathcal{P}_i < \tau_{\mathrm{mask}}\}.
    \label{eq:mask_prune}
\end{equation}
\noindent
We further suppress unsupported densification by applying a mask-support gating function to the accumulated positional gradient used for clone and split operations:
$\tilde{\mathbf{g}}_i = \mathbf{g}_i \cdot \mathcal{S}_v(\mathbf{x}_i)$.
Here, $\mathcal{S}_v(\mathbf{x}_i)=\mathbf{1}[M_v(\pi_v(\mathbf{x}_i))>0]$ is a binary support indicator that becomes one when Gaussian $g_i$ is projected inside the object mask in view $v$, and zero otherwise.
Thus, Gaussians projected outside the object mask do not trigger densification.

\subsubsection{Appearance-guided Local Gaussian Merging.}
 
While mask-guided pruning removes background artifacts, redundant and overlapping Gaussians can still accumulate on the foreground object. To improve object-centric reconstruction fidelity, we introduce an appearance-guided local Gaussian merging strategy that consolidates nearby primitives with similar appearance. For each Gaussian, neighboring primitives are identified within a spatial radius proportional to its scale.
Among these neighboring primitives, Gaussians with similar color are selected, and merging is performed when the number of candidates exceeds a predefined threshold.
The merged Gaussian replaces the group by averaging the geometric and appearance attributes—position, scale, rotation, and color—while accumulating opacity values to preserve the overall density of the region.

\subsection{Silhouette-aligned Object Loss}
\label{sec:oc_loss}
Existing object-level methods often use boundary or occupancy constraints to refine object extraction after scene-level reconstruction. In contrast, our objective is designed for direct object reconstruction from masked observations. 
Even with object-masked supervision, photometric losses do not explicitly align the rendered Gaussian support with the object silhouette, which allows background leakage or incomplete object support to persist. We therefore introduce a silhouette-aligned object loss that combines object-focused photometric reconstruction with silhouette-level support alignment.
Given an input image $I$, a binary object mask $M$, and the masked target image $\tilde{I} = I \odot M$, we define the object-weighted photometric loss as
\begin{equation}
\mathcal{L}_{\mathrm{obj}} =
\frac{1}{N} \sum_{p=1}^{N}
\left[
M_p\,|\hat{I}_p - \tilde{I}_p|
+
\gamma(1-M_p)\,|\hat{I}_p - \tilde{I}_p|
\right].
\label{eq:mask_l1}
\end{equation}
where $\hat{I}$ is the rendered image and $\gamma$ is a small attenuation factor for background pixels. 
This loss emphasizes reconstruction inside the object mask while weakly penalizing residual colors outside the mask.
To further align the spatial support of Gaussians with the target object, we render a soft silhouette $\hat{M}$ from the current Gaussian representation and define the silhouette loss as
\begin{equation}
\mathcal{L}_{\mathrm{sil}} =
1 -
\frac{
\sum_{p=1}^{N} \left(\hat{M}_p M_p\right) + \epsilon
}{
\sum_{p=1}^{N} \left(\hat{M}_p + M_p - \hat{M}_p M_p\right) + \epsilon
}.
\label{eq:sil_loss}
\end{equation}
The final objective is
\begin{equation}
    \mathcal{L}_{\mathrm{final}}
    =
    (1-\lambda_{\mathrm{dssim}})\mathcal{L}_{\mathrm{obj}}
    +
    \lambda_{\mathrm{dssim}}
    \left(1-\mathrm{SSIM}(\hat{I}, \tilde{I})\right)
    +
    \lambda_{\mathrm{sil}}\mathcal{L}_{\mathrm{sil}} .
    \label{eq:loss_final}
\end{equation}
This objective aligns appearance reconstruction and silhouette support with the target object. It encourages the Gaussian distribution to remain compact and object-centric during direct reconstruction.

\section{Experiments}
\label{sec:experiments}

\subsection{Implementation Details}

Our method follows the default 3DGS\cite{3dgs} configuration, except that the densification gradient threshold is increased from 0.0002 to 0.0003. For datasets requiring generated masks, we use EfficientViT-SAM\cite{zhang2024efficientvitsam} as the segmentation model, but our method is not tied to a specific mask generator; additional results with other mask generators are provided in the supplementary material. All experiments are conducted on a single NVIDIA A6000 GPU and ablation studies are performed on a single NVIDIA RTX 4070 Ti Super GPU. Further details on the experiments, including hit-ratio threshold settings, filtering-schedule sensitivity, and mean and standard deviation over three seeds for the key comparisons in Tables~\ref{tab:quantitative_results} and~\ref{tab:object_difficulty}, are provided in the supplementary material.

\begin{table*}[!t]
\centering
\caption{Quantitative comparison on the LERF-Mask~\cite{Ye2024GaussianGrouping}, Mip-NeRF 360~\cite{mipnerf360}, and Tanks and Temples~\cite{knapitsch2017tanks}. Results are averaged over all objects, with all methods evaluated under the same controlled masked-input protocol. The percentages next to each dataset indicate the average object area ratio within the frame. The best, second-best, and third-best results are highlighted in red, orange, and yellow, respectively.}
\label{tab:quantitative_results}
\footnotesize
\setlength{\tabcolsep}{10pt}
\setlength{\aboverulesep}{0pt} 
\setlength{\belowrulesep}{0pt} 
\renewcommand{\arraystretch}{0.97} 
\resizebox{0.8\linewidth}{!}{
    \begin{tabular}{c | l | c c c r }
    \toprule
    \textbf{Category} & \textbf{Method} & PSNR $\uparrow$ & SSIM $\uparrow$ & LPIPS $\downarrow$ & \begin{tabular}[c]{@{}c@{}}FPS$\uparrow$\end{tabular}  \\
    \midrule
    \multicolumn{6}{c}{\cellcolor{gray!12}\textbf{LERF-Mask} (2.01\%)} \\
    \midrule
    Baseline & 3DGS~\cite{3dgs}          & \cellcolor{yellow!20}40.67 & \cellcolor{yellow!20} 0.990 &\cellcolor{yellow!20} 0.010 &  639   \\
    \cmidrule{1-2}
    Efficient GS & PUP 3D-GS~\cite{pup}      & 39.50 &  0.989 &  0.012 & \cellcolor{orange!20} 1,934   \\
    \cmidrule{1-2}
    \multirow{2}{*}{Object-centric} & ObjectGS~\cite{Zhu2025ObjectGS}      & 37.65 & 0.988 & 0.013 & 350 \\
    \cmidrule{2-2}
     & Trace3D~\cite{shen2025trace3d}      & 34.74 & 0.986 & 0.019 & \cellcolor{yellow!20}772 \\
    \cmidrule{1-2}
    \multirow{1}{*}{High-Fidelity} & FDS-GS~\cite{zeng2025frequency}        & \cellcolor{orange!20} 40.84 & \cellcolor{orange!20} 0.990 & \cellcolor{orange!20} 0.010 & 449 \\
    \cmidrule{1-6}
    & \textbf{Ours} & \cellcolor{red!20} 42.89 & \cellcolor{red!20} 0.992 & \cellcolor{red!20} 0.008 &  \cellcolor{red!20} 2,081   \\
    \midrule
    \midrule
    \multicolumn{6}{c}{\cellcolor{gray!12}\textbf{Mip-NeRF 360} (12.54\%)} \\
    \midrule
    Baseline & 3DGS~\cite{3dgs}          & \cellcolor{yellow!20} 32.99 & \cellcolor{yellow!20} 0.977 & \cellcolor{orange!20} 0.023 &  250 \\
    \cmidrule{1-2}
     Efficient GS & PUP 3D-GS~\cite{pup}      & 32.67 & 0.975 & 0.028 & \cellcolor{red!20} 1,121 \\
    \cmidrule{1-2}
    \multirow{2}{*}{Object-centric} & ObjectGS~\cite{Zhu2025ObjectGS}      & 29.22 & 0.958 & 0.050 & 177  \\
    \cmidrule{2-2}
     & Trace3D~\cite{shen2025trace3d}      & 31.20 &0.971 & 0.032 & \cellcolor{yellow!20}405 \\
    \cmidrule{1-2}
    \multirow{1}{*}{High-Fidelity} & FDS-GS~\cite{zeng2025frequency}        & \cellcolor{orange!20} 33.46 & \cellcolor{red!20} 0.979 & \cellcolor{red!20} 0.022 &   385  \\
    \cmidrule{1-6}
    & \textbf{Ours} & \cellcolor{red!20} 34.49 & \cellcolor{orange!20} 0.977 & \cellcolor{yellow!20} 0.024 & \cellcolor{orange!20}  1,010  \\
   \midrule
   \midrule
    \multicolumn{6}{c}{\cellcolor{gray!12}\textbf{Tanks \& Temples} (31.35\%)} \\
    \midrule
    Baseline & 3DGS~\cite{3dgs}          & \cellcolor{orange!20} 30.63 & \cellcolor{orange!20} 0.964 & \cellcolor{orange!20} 0.048 & 159   \\
    \cmidrule{1-2}
     Efficient GS & PUP 3D-GS~\cite{pup}      & 29.98 & 0.955 & 0.068 & \cellcolor{red!20} 735   \\
    \cmidrule{1-2}
    \multirow{2}{*}{Object-centric} & ObjectGS~\cite{Zhu2025ObjectGS}      & 30.21 & 0.960 & 0.056 & 123  \\
    \cmidrule{2-2}
     & Trace3D~\cite{shen2025trace3d}      & 26.33 & 0.923 & 0.103 & \cellcolor{orange!20}508 \\
    \cmidrule{1-2}
    \multirow{1}{*}{High-Fidelity} & FDS-GS~\cite{zeng2025frequency}        & \cellcolor{yellow!20} 30.25 & \cellcolor{red!20} 0.965 & \cellcolor{red!20} 0.041 &  231  \\
    \cmidrule{1-6}
    & \textbf{Ours} & \cellcolor{red!20} 30.94 & \cellcolor{yellow!20} 0.961 & \cellcolor{yellow!20} 0.056 &  \cellcolor{yellow!20} 397 \\
    \bottomrule
    \end{tabular}
}
\end{table*}

\subsection{Rendered Results of Object-centric 3D Gaussian Splatting}

\label{sec:standard_eval}
\begin{figure*}[!t]
    \centering
    \includegraphics[width=0.7\textwidth]{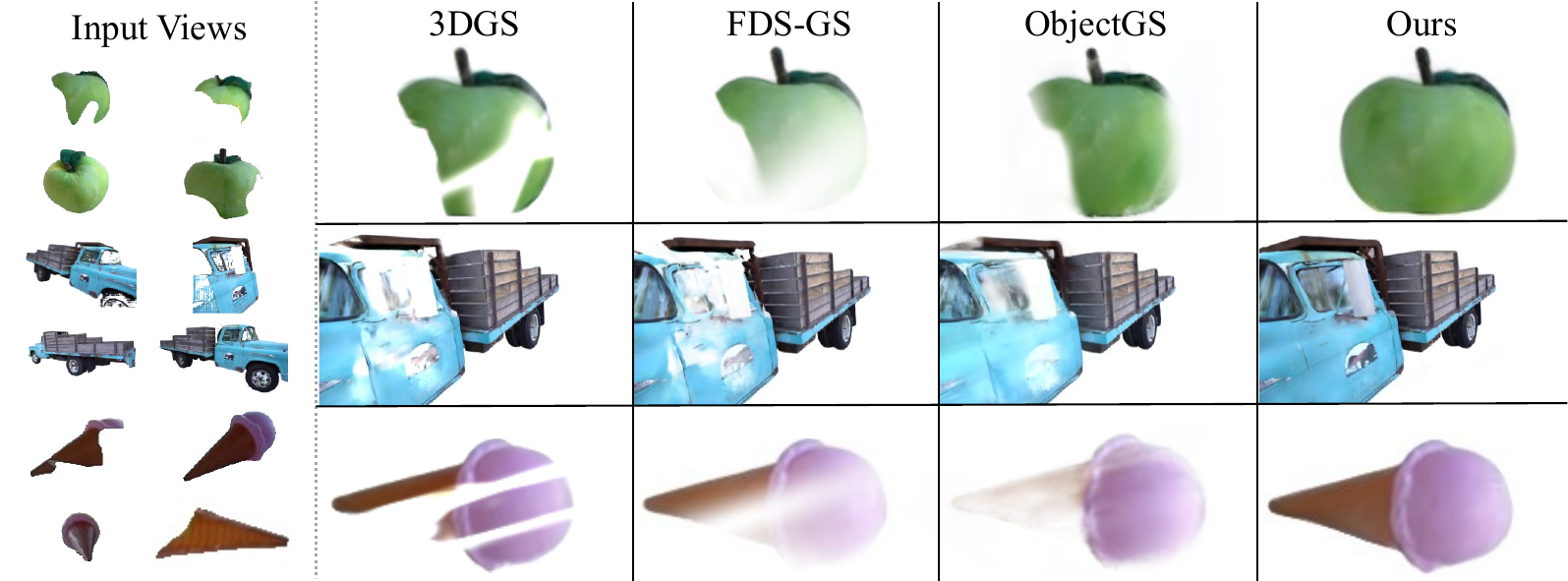}
    \caption{MaRO-GS  reconstructs target objects from input views with inconsistent background removal. Existing approaches produce residual artifacts, whereas MaRO-GS  achieves robust, high-fidelity object reconstruction.}
    \label{fig:teaser}
\end{figure*}

\Cref{tab:quantitative_results} reports quantitative comparisons on LERF-Mask\cite{Ye2024GaussianGrouping}, Mip-NeRF 360\cite{mipnerf360}, and Tanks and Temples\cite{knapitsch2017tanks}. All methods are evaluated under the same object-centric masked-input protocol without method-specific modifications, and the metrics measure agreement with the mask-defined target $\tilde{I}=I\odot M$ rather than absolute object appearance. Additional comparisons under the baselines' native full-scene settings are provided in the supplementary material. Our method achieves the highest PSNR on all three datasets. The performance gap widens as the object region occupies a smaller portion of the image, with the most pronounced improvement on LERF-Mask (2.01\%), where our method outperforms the second-best method by 2.05 dB. Qualitative comparisons are provided in \cref{fig:teaser}. Our method produces cleaner object reconstructions with fewer background artifacts while more reliably recovering object regions occluded by other objects.

\begin{figure}[t]
     \centering
        \includegraphics[width=0.9\linewidth]{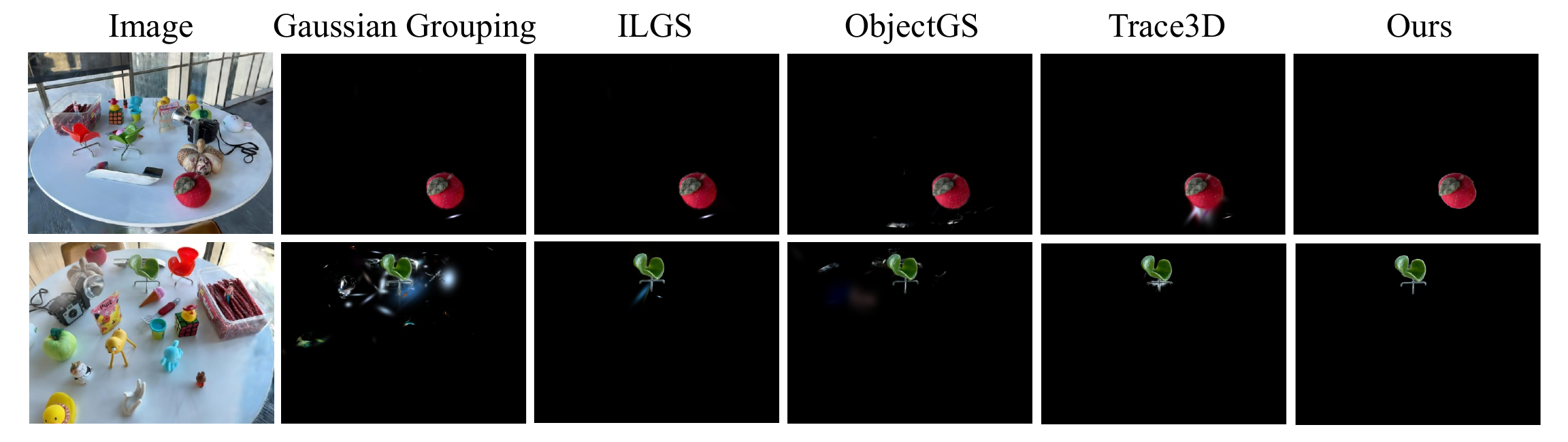}
        \captionof{figure}{
        Qualitative comparison of our method with object extraction methods on the LERF-Mask dataset.
        }
        \label{fig:extraction}
\end{figure}

\subsection{Evaluation of Object Boundary Quality}
\label{sec:extracted_eval}
We evaluate the boundary quality of target objects on the LERF-Mask dataset. 
This evaluation is conducted on images that include background regions to assess object extraction quality in the original scene context. 
For existing object-level methods, we evaluate the boundaries of extracted objects. 
For MaRO-GS, which directly reconstructs target objects, we evaluate the boundaries of the reconstructed objects. 
Both evaluations follow the protocol of FlashSplat~\cite{shen2024flashsplat} and BEA-GS~\cite{mazzucchelli2026bea}. 
\Cref{tab:segmentation_results} reports quantitative comparisons in terms of mIoU and bIoU, and \cref{fig:extraction} provides qualitative comparisons of target object boundaries. 
Our method achieves the best performance, with 86.24 mIoU and 83.04 bIoU, outperforming Trace3D by 2.66 and 3.36 points, respectively. 
These results demonstrate that our method produces a more accurate object representation with fewer boundary errors and less background leakage.

\subsection{Additional Evaluation for Object-centric Reconstruction} \label{sec:dataset}

Existing 3DGS benchmarks are primarily designed for reconstruction under consistent supervision and therefore cannot adequately measure robustness under multi-view inconsistencies.
Moreover, conventional metrics may not capture perceptually meaningful visual quality differences when the target object occupies only a small image region, as shown in \cref{fig:oc_problem}.
To address these limitations, we conduct two additional evaluations for object-centric scenarios: (1) \textit{Robustness Evaluation under Mask Errors}, which measures reconstruction robustness under varying levels of mask inconsistency, and (2) \textit{Object-centric Viewpoint Evaluation}, which assesses perceptual fidelity from object-centric viewpoints.

\begin{figure*}[!t]
    \centering

    \begin{minipage}[t]{0.35\textwidth}
        
        \centering
        \captionof{table}{
        Quantitative comparison of object boundary quality. Best results are highlighted in \textbf{bold}.
        }
        \label{tab:segmentation_results}

        \resizebox{1.0\linewidth}{!}{%
        \begin{tabular}{lcc r}
        \toprule
        \textbf{Method} & \textbf{mIoU} $\uparrow$ & \textbf{bIoU} $\uparrow$ & \textbf{FPS} $\uparrow$ \\
        \midrule
        GaussGroup.\cite{Ye2024GaussianGrouping}  & 30.63 & 27.39 & 857 \\
        ILGS\cite{jang2025identity}         & 45.40 & 36.54 & 762 \\
        ObjectGS\cite{Zhu2025ObjectGS}     & 76.78 & 74.42 & 70 \\
        Trace3D\cite{shen2025trace3d}      & 83.58 & 79.68 & 627 \\
        \midrule
        \textbf{Ours} & \textbf{86.24} & \textbf{83.04} & \textbf{1,106} \\
        \bottomrule
        \end{tabular}
        }
    \end{minipage}
    \hfill
    \begin{minipage}[t]{0.61\textwidth}
 \centering
        \captionof{table}{Evaluation on the DTU~\cite{dtu} Dataset. Best results are highlighted in \textbf{bold}, and second-best results are \underline{underlined}. PUP 3D-GS~\cite{pup} fails to train because all Gaussians are pruned during training.}
        \label{tab:object_difficulty}
        \setlength{\tabcolsep}{0.2pt}
        \resizebox{1.0 \columnwidth}{!}{%
        \begin{tabular}{l | ccc | ccc | ccc}
        \toprule
        \multirow{2}{*}{\textbf{Method}} 
        & \multicolumn{3}{c}{\textbf{Easy}} 
        & \multicolumn{3}{c}{\textbf{Medium}} 
        & \multicolumn{3}{c}{\textbf{Hard}} \\
        \cmidrule(lr){2-4} \cmidrule(lr){5-7} \cmidrule(lr){8-10}
        & PSNR$\uparrow$ & SSIM$\uparrow$ & LPIPS$\downarrow$
        & PSNR$\uparrow$ & SSIM$\uparrow$ & LPIPS$\downarrow$
        & PSNR$\uparrow$ & SSIM$\uparrow$ & LPIPS$\downarrow$ \\
        \midrule
        
        3DGS\cite{3dgs} & 31.15 & 0.955 & \underline{0.055} & 22.52 & 0.895 & 0.116 & 19.60 & 0.906 & \underline{0.106} \\
        ObjectGS\cite{Zhu2025ObjectGS}
        & \underline{31.51} & \textbf{0.968} & 0.056 
        &\underline{23.90} & \underline{0.919} & \underline{0.111} 
        & \underline{21.41} & \underline{0.911} & \underline{0.106} \\
        PUP 3D-GS\cite{pup} &  & ERROR &  &  & ERROR &  &  & ERROR &  \\
        FDS-GS\cite{zeng2025frequency}
        & 31.04 & \underline{0.964} & \textbf{0.053} 
        & 22.16 & 0.892 & 0.117 
        & 19.15 & 0.909 & \underline{0.106} \\
        \midrule
        \textbf{Ours} 
        & \textbf{33.26} & 0.963 & \underline{0.055}
        & \textbf{26.50} & \textbf{0.939} & \textbf{0.085} 
        & \textbf{22.88} & \textbf{0.924} & \textbf{0.085} \\
        
        \bottomrule
        \end{tabular}
        }
      
    \end{minipage}
\end{figure*}

\begin{figure*}[!t]
    \centering
    \includegraphics[width=0.6\textwidth]{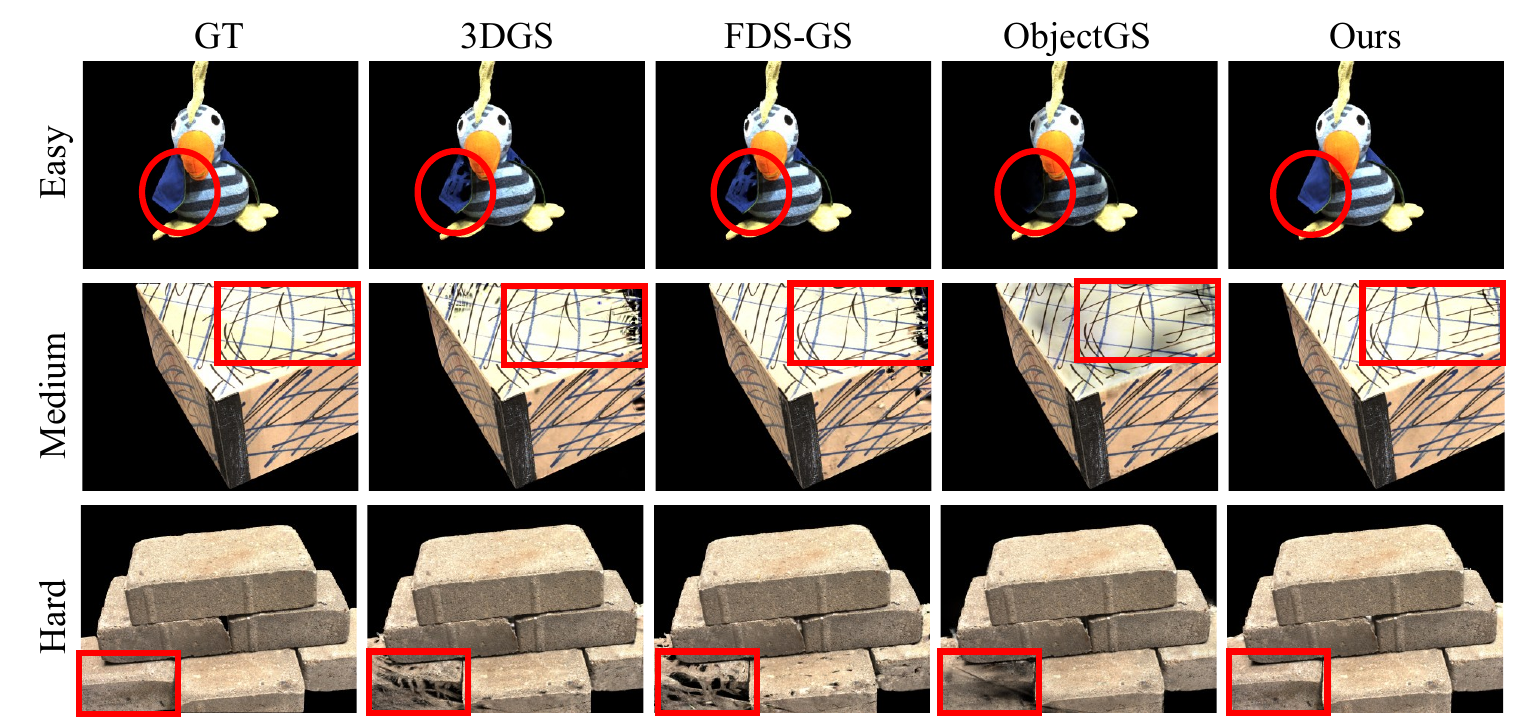}
    \caption{Qualitative comparisons on the DTU~\cite{dtu} Dataset. Our method remains more robust and preserves the object geometry more reliably despite mask inaccuracies. }
    \label{fig:mask_noise}
\end{figure*}

\subsubsection{Robustness Evaluation under Mask Errors.} \label{sec:dtu}
To evaluate robustness under imperfect object-mask supervision, we construct a controlled evaluation using DTU~\cite{dtu}. 
We divide samples into Easy, Medium, and Hard groups according to the mask error ratio: at most 10\%, above 10\% up to 50\%, and above 50\%, respectively.
DTU uses EfficientViT-SAM-predicted masks without synthetic error injection, and the mask-error ratio denotes the fraction of manually identified erroneous views per scene. Further details are provided in the supplementary material.
\Cref{tab:object_difficulty} reports quantitative results on the DTU subset. 
Baseline methods degrade as mask errors become more severe, and PUP 3D-GS fails to train because all Gaussians are pruned under inconsistent inputs. 
In contrast, MaRO-GS  achieves the best PSNR across all difficulty levels, with 33.26 dB, 26.50 dB, and 22.88 dB on Easy, Medium, and Hard cases, respectively. 
Additional DTU filtering ablations in the supplementary material show that view filtering is most effective on Medium, where corrupted and reliable views coexist, while it provides no additional gain on Hard, where most views are corrupted. As shown in \cref{fig:mask_noise}, MaRO-GS  preserves the target object structure more reliably than competing methods even under significant segmentation noise.

\subsubsection{Object-centric Viewpoint Evaluation.} \label{sec:oc_eval}
\begin{figure*}[!t]
    \centering
    \begin{minipage}[t]{0.48\textwidth}
        \vspace{0pt}
        \centering
        \includegraphics[width=1.0\linewidth]{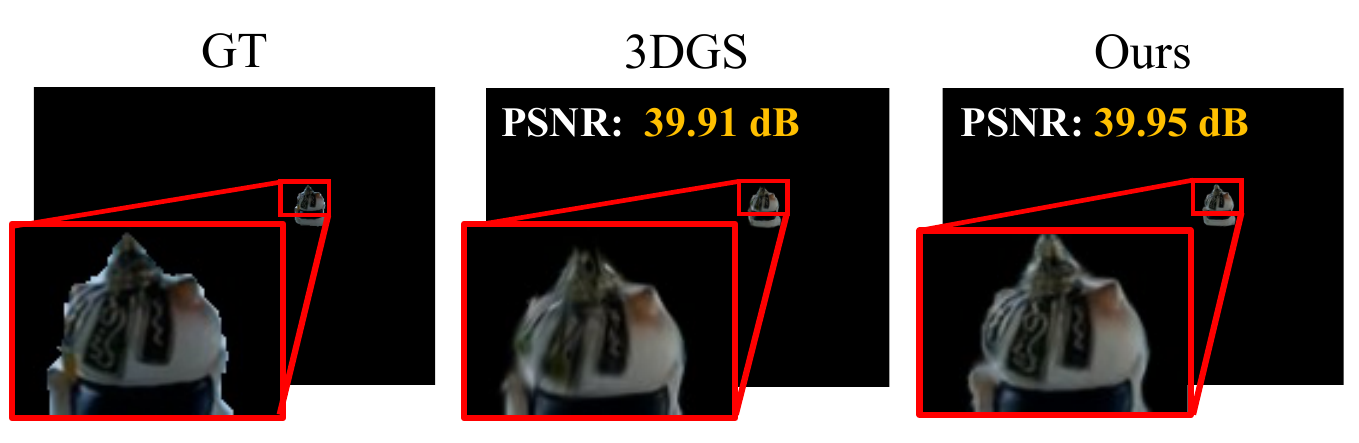} 
        \caption{Limitation of standard 3DGS evaluation. Although the reconstructed object quality differs noticeably, the PSNR values remain nearly identical.}
        \label{fig:oc_problem}
    \end{minipage}
    \hfill
    \begin{minipage}[t]{0.48\textwidth}
    
        \centering
        \captionof{table}{Results of Object-centric evaluation on the LERF-Mask dataset. Best results are highlighted in \textbf{bold}.}
        \label{tab:nr_metrics}
        \setlength{\tabcolsep}{1pt}
        \resizebox{1.0\linewidth}{!}{
            \begin{tabular}{l | c c | c}
            \toprule
            \textbf{Method} 
            & \begin{tabular}{@{}c@{}}\textbf{CLIP-IQA+}$\uparrow$ \\ \cite{wang2023exploring}\end{tabular}
            & \begin{tabular}{@{}c@{}}\textbf{MUSIQ}$\uparrow$ \\ \cite{ke2021musiq}\end{tabular}
            & \begin{tabular}{@{}c@{}}\textbf{Met3R}$\downarrow$ \\ \cite{asim2025met3r}\end{tabular} \\
            \midrule
            ObjectGS\cite{Zhu2025ObjectGS}  & 0.464 & 38.85 & 0.144 \\
            Trace3D\cite{shen2025trace3d}  & 0.469 & 48.06 & 0.088 \\
            \midrule
            \textbf{Ours} &\textbf{ 0.489 }& \textbf{55.64} & \textbf{0.073} \\
            \bottomrule
            \end{tabular}%
        }
    
    \end{minipage}
\end{figure*}

When the target object occupies only a small portion of the image, full-image metrics become insensitive to perceptually meaningful differences in the object region, as shown in \cref{fig:oc_problem}. 
To complement the standard evaluation, we additionally evaluate objects from object-centric viewpoints, where novel views are rendered around the object.
Since the transformed viewpoints do not correspond to available ground-truth images, perceptual quality is assessed using no-reference metrics: CLIP-IQA+~\cite{wang2023exploring}, MUSIQ~\cite{ke2021musiq} and Met3R~\cite{asim2025met3r}. As shown in \Cref{tab:nr_metrics}, our method achieves the best results across all metrics. Details of the evaluation pipeline are provided in the supplementary material.

\begin{figure*}[!t]
    \centering
    \begin{minipage}[t]{0.48\textwidth}
        
        \centering
        \captionof{table}{Ablation study on the LERF-Mask~\cite{Ye2024GaussianGrouping} dataset. (a)~mask-guided pruning and densification, (b)~appearance-guided local Gaussian merging, (c)~geometric view consistency, (d)~region-based view validation.}
        \label{tab:ablation_full}
        \setlength{\tabcolsep}{0.5pt}
        \resizebox{0.9\linewidth}{!}{%
        \begin{tabular}{cccc|cccc}
        \toprule
        \multicolumn{4}{c|}{\textbf{Component}} & \multicolumn{4}{c}{\textbf{Performance}} \\
        \cmidrule(r){1-4}\cmidrule(l){5-8}
        \textbf{(a)} & \textbf{(b)} & \textbf{(c)} & \textbf{(d)}
          & \textbf{PSNR\,$\uparrow$}
          & \begin{tabular}[c]{@{}c@{}}\textbf{Train.}\\\textbf{Time (s)$\downarrow$}\end{tabular}
          & \begin{tabular}[c]{@{}c@{}}\textbf{FPS}\textbf{$\uparrow$}\end{tabular}
          & \textbf{\#Gauss.\,$\downarrow$} \\
        \midrule
                   &            &            &            & 41.98  & 464.69 & 1,458 & 42,846 \\
        \checkmark &            &            &            & 42.54  & 340.92 & 1,630 & 25,805 \\

        \checkmark & \checkmark &            &            & 42.51  & 332.00 & 1,784 & 21,157 \\
        \checkmark & \checkmark & \checkmark &            & 42.55  & 340.07 & 1,757 & 25,540 \\

        \midrule
        \checkmark & \checkmark & \checkmark & \checkmark & 42.90 & 322.84 & 1,653 & 24,422 \\
        \bottomrule
        \end{tabular}
        }

    \end{minipage}
    \hfill
    \begin{minipage}[t]{0.48\textwidth}

        \centering
        \setlength{\tabcolsep}{3pt}
        \captionof{table}{Ablation study on the LERF-Mask~\cite{Ye2024GaussianGrouping} dataset on the effect of object loss and silhouette loss. Best results are highlighted in \textbf{bold}, and second-best results are \underline{underlined}.}
        \label{tab:ablation}
        \begin{tabular}{cc|ccc}
        \midrule
        $\mathcal{L}_{obj}$ & $\mathcal{L}_{sil}$ & PSNR$\uparrow$ & mIoU$\uparrow$ & bIoU$\uparrow$ \\
        \midrule
         &  & 39.25 & 27.77 & 23.72 \\
         \checkmark &  & \textbf{41.92} & 47.37 & 41.16 \\
         &  \checkmark & 39.20 & \underline{85.69} & \textbf{82.85} \\
         \midrule
         \checkmark &  \checkmark & \underline{41.51} & \textbf{85.76} & \underline{82.45} \\
        \bottomrule
        \end{tabular}

    \end{minipage}
\end{figure*}

\section{Ablation Study}
\label{sec:ablation}

\subsection{Ablations on Each Module}
\Cref{tab:ablation_full} presents an ablation study of the proposed components on the LERF-Mask dataset~\cite{Ye2024GaussianGrouping}. All variants in this table are trained with the proposed silhouette-aligned object loss. Applying \textit{Mask-guided Pruning and Densification} (a) improves PSNR and reduces the number of Gaussians, which indicates that removing background-supported Gaussians leads to a more compact object representation. Adding \textit{appearance-guided local Gaussian merging} (b) further reduces the Gaussian count and improves FPS with negligible quality degradation. With \textit{geometric view consistency} (c), the model maintains comparable quality while improving supervision stability. 
Finally, adding \textit{region-based view validation} (d) achieves the best PSNR. \Cref{tab:ablation} analyzes the effect of the proposed silhouette-aligned object loss. Without both, the model produces poor object extraction quality. Using only $\mathcal{L}_{obj}$ achieves the highest PSNR, but its mIoU and bIoU remain limited. Using only $\mathcal{L}_{\mathrm{sil}}$ greatly improves mIoU and bIoU, confirming that silhouette supervision effectively reduces background leakage and boundary errors. Combining $\mathcal{L}_{\mathrm{obj}}$ and $\mathcal{L}_{\mathrm{sil}}$ achieves the best mIoU while maintaining boundary quality.

\subsection{Effect of Object-supported Gaussian Density Control}
\begin{figure*}[!t]
    \centering
    \begin{minipage}[t]{0.43\textwidth}
    \centering
    \includegraphics[width=1.0\linewidth]{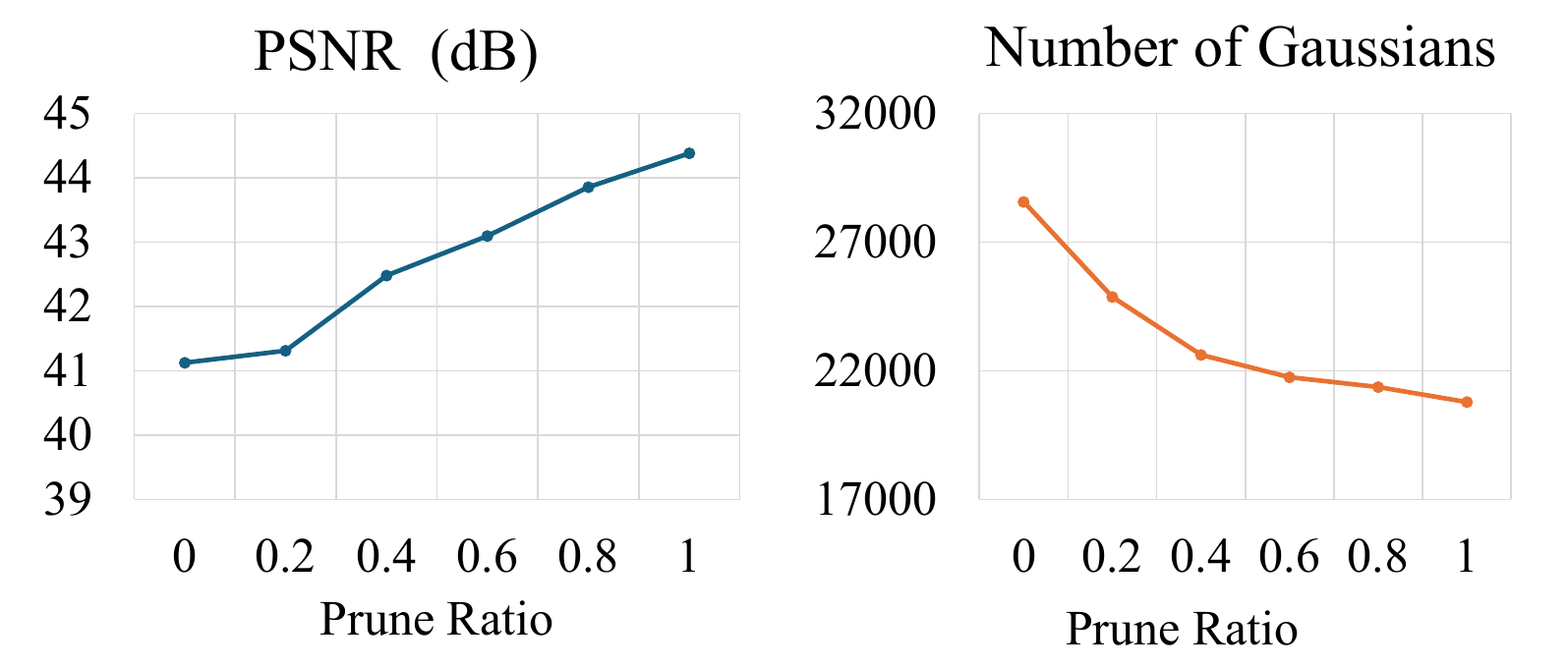}
    \caption{Effect of pruning ratio.}
    \label{fig:prune_ratio}
    \end{minipage}
    \hfill
    \begin{minipage}[t]{0.51\textwidth}
    \centering
    \includegraphics[width=0.9\linewidth]{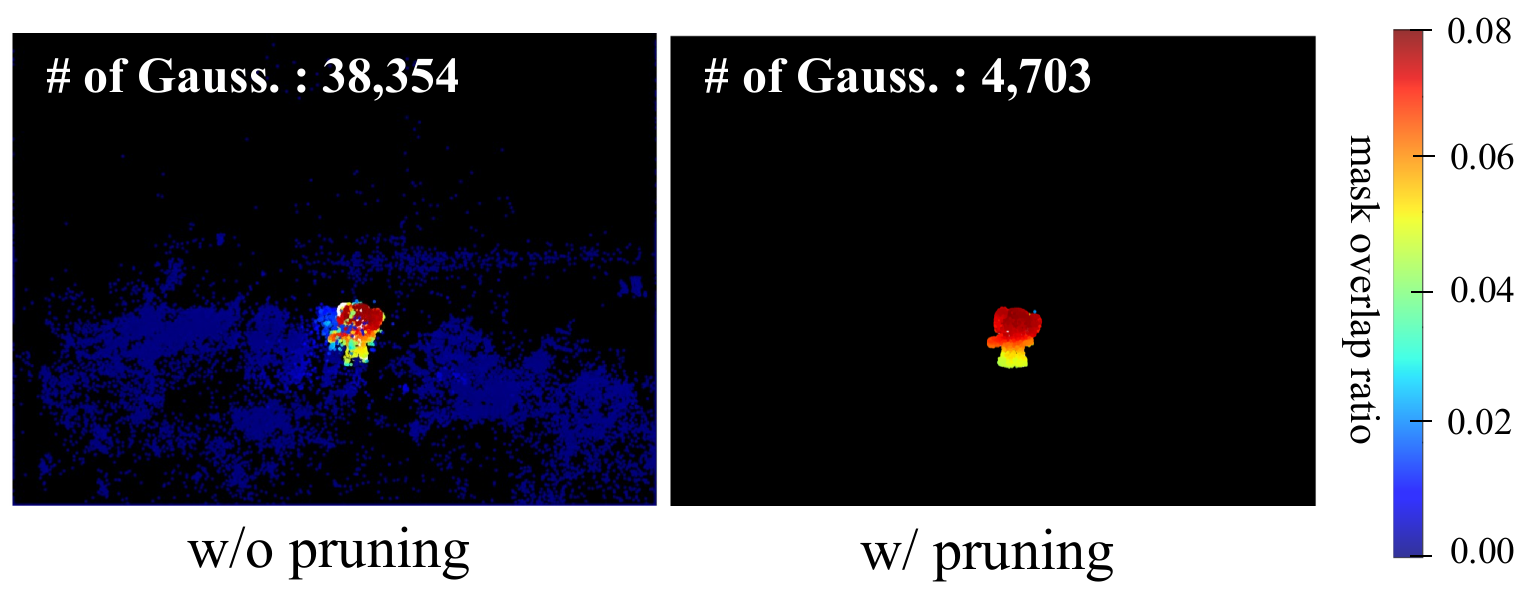}
    \caption{Visualization of Gaussian centers without and with applying mask-guided pruning.}
    
    \label{fig:pruning_vis}
    \end{minipage}
\end{figure*}

We analyze the effect of different pruning ratios in the proposed object-supported Gaussian density control. As shown in \cref{fig:prune_ratio}, increasing the pruning ratio significantly reduces the number of Gaussians. Moreover, PSNR consistently improves, suggesting that removing redundant background Gaussians helps the model focus on object regions.
\cref{fig:pruning_vis} visualizes the spatial distribution of Gaussian centers with and without the proposed pruning. Without pruning, a large number of Gaussians are scattered outside the target object region. With pruning, Gaussians are concentrated within the object boundary, which yields a more compact and accurate representation.

\subsection{Efficiency of Object-centric Reconstruction}
\Cref{tab:runtime} compares the computational efficiency of different methods on LERF-Mask~\cite{Ye2024GaussianGrouping}. All methods use the same object-masked multi-view observations. MaRO-GS achieves substantially lower overall runtime and memory consumption than ObjectGS while maintaining computational cost comparable to standard 3DGS.

\begin{table}[!t]
\centering
\caption{Runtime comparison of different methods on LERF-Mask dataset.}
\label{tab:runtime}
\resizebox{0.7\linewidth}{!}{
    \setlength{\tabcolsep}{3pt}
    \begin{tabular}{l | c c c c | c}
    \toprule
    \multirow{2}{*}{Method} & Preprocessing & Training & Rendering & Total & Max VRAM \\
     & Time (s) & Time (s) & Time (s) & Time (s) & (MB) \\
    \midrule
    3DGS~\cite{3dgs}        & N/A & 363  & 4.84 & 367.84  & 4,305 \\
    PUP 3D-GS~\cite{pup}    & 363 & 103  & 4.69 & 470.69  & 3,551 \\
    ObjectGS~\cite{Zhu2025ObjectGS}    & 338 & 2,196 & 8.07 & 2,542.07 & 24,300 \\
    \midrule
    Ours        & N/A & 345 & 3.53 & 348.53  & 4,187 \\
    \bottomrule
    \end{tabular}
}
\end{table}

\section{Limitations}

Although MaRO-GS improves robustness under inconsistent mask supervision, severe segmentation failures can still degrade reconstruction quality. The Medium and Hard DTU subsets are limited to 3 and 4 scenes, respectively, and conservative filtering achieves only 4.6\% recall on Hard, so the overall Hard-case improvement cannot be attributed to view filtering alone. Filtering also depends on sufficiently localized Gaussian support during optimization. Further analyses and failure cases are provided in the supplementary material.

\section{Conclusion}
\label{sec:conclusion}
We present MaRO-GS for object-centric 3D Gaussian Splatting under inconsistent mask supervision. MaRO-GS combines mask-reliability view filtering with object-supported Gaussian density control to exclude unreliable supervision and suppress Gaussians irrelevant to the target object. Experiments demonstrate improved PSNR and robustness, complemented by object-centric and mask-error robustness evaluations.

\noindent\textbf{Acknowledgments.}
This work was partly supported by the National Research Foundation of Korea(NRF) grant funded by the Korea government(MSIT)(RS-2025-00520308, 50\%) and the Glocal 30 program of Kyungpook National University, funded by the Ministry of Education(25\%) and by Culture, Sports and Tourism R\&D Program through the Korea Creative Content Agency grant funded by the Ministry of Culture, Sports and Tourism in 2026(RS-2026-2552393, 25\%).

\bibliographystyle{splncs04}
\bibliography{main}
\clearpage
\appendix

{\centering\Large\bfseries Supplementary Material\par}
\vspace{1em}

\section*{Overview}

This supplementary material provides additional details and analyses that support the main paper. 
\cref{sec:A_Implementation_details} summarizes the implementation details of MaRO-GS, including the optimization settings and module-specific hyperparameters. 
\cref{sec:B_0} describes the experimental protocols used throughout the paper, covering dataset configurations, mask generation and annotation procedures, object-region evaluation, boundary quality evaluation, robustness analysis under mask errors, and object-centric viewpoint evaluation. 
\cref{sec:C_0} presents additional experiments and analyses, including comparisons across evaluation settings, additional ablation studies, analyses of mask-reliability view filtering and object-supported Gaussian density control, and sensitivity studies of key hyperparameters.
\cref{sec:D_0} analyzes practical challenges in object-centric reconstruction, including mask errors from automatic segmentation and artifacts caused by background removal.
\cref{sec:E_0} clarifies the relationship between MaRO-GS and closely related methods by comparing their task settings, supervision strategies, pruning mechanisms, and use of Gaussian visibility. 
\cref{sec:F_0} discusses remaining limitations and representative failure cases, both for MaRO-GS and for existing baseline methods. 
\cref{sec:G_0} evaluates whether MaRO-GS remains robust when masks are generated by different segmentation models. 
\cref{supple_seed} reports seed variance for the key comparisons in the main paper.
Finally, \cref{sec:H_0} provides additional qualitative reconstruction results on LERF-Mask, Mip-NeRF 360, and Tanks and Temples.

\section{Implementation Details}

\label{sec:A_Implementation_details}

We follow the default 3DGS optimization settings unless otherwise specified. All models are trained for 30K iterations with spherical harmonics degree 3. We use the default learning rates for Gaussian position, feature, opacity, scaling, and rotation parameters, and set the DSSIM weight to $\lambda_{\mathrm{dssim}}=0.2$. For the \textit{silhouette-aligned object loss}, we set the background weight to $\gamma=0.01$ in all experiments.
For the proposed modules, we use the following configuration in all experiments. 
Mask-reliability view filtering is enabled by default. 
\textit{Geometric view consistency }is performed from iteration 1K to 5K at 2K-iteration intervals, using up to 20 pose-neighboring views selected by camera-center distance and the top-3 weighted similarities for each view score. 
The adaptive largest-gap threshold is accepted only when the low-score cluster occupies at most 25\% of the views and the cluster separation is larger than 0.5. 
\textit{Region-based view validation} is applied once at iteration 1.8K with a hit-ratio threshold of 0.05. 
For \textit{object-supported Gaussian density control}, mask-guided pruning is performed at iterations 1K, 3K, and 5K with pruning coefficient $k=0.5$, maximum pruning threshold 0.05, and prune ratio 1.0. 
\textit{Appearance-guided local Gaussian merging} is applied at scheduled filtering iterations during the early optimization stage. Specifically, merging is triggered when the iteration is divisible by the filtering interval and remains below the filtering end iteration. With the default setting of a filtering interval of 2K and a filtering end iteration of 5K, we use a color threshold of 0.2, a radius factor of 0.4, and a minimum group size of 3.

\section{Experimental Details}
\label{sec:B_0}
\subsection{Rendered Results of Object-centric 3D Gaussian Splatting \\ (Sec 4.2) }

\label{sec:B_1}

\subsubsection{Datasets and Mask Annotation}

\begin{figure}[t]
    \centering
    \includegraphics[width=0.9\linewidth]{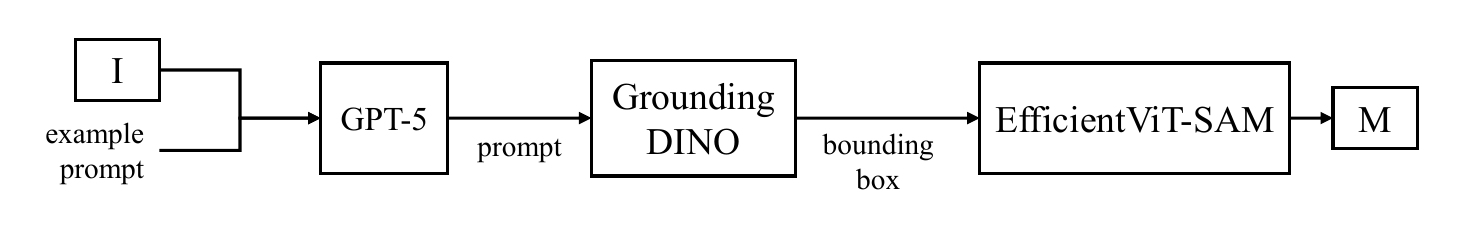}
    \caption{Overview of the mask generation pipelines. GPT-5~\cite{openai2025gpt5} generates a text prompt, GroundingDINO~\cite{liu2024grounding} predicts a bounding box, and EfficientViT-SAM produces the object mask.}
    \label{fig:mask_pipeline}
\end{figure}

\begin{figure}[!t]
    \centering
    \includegraphics[width=0.95\linewidth]{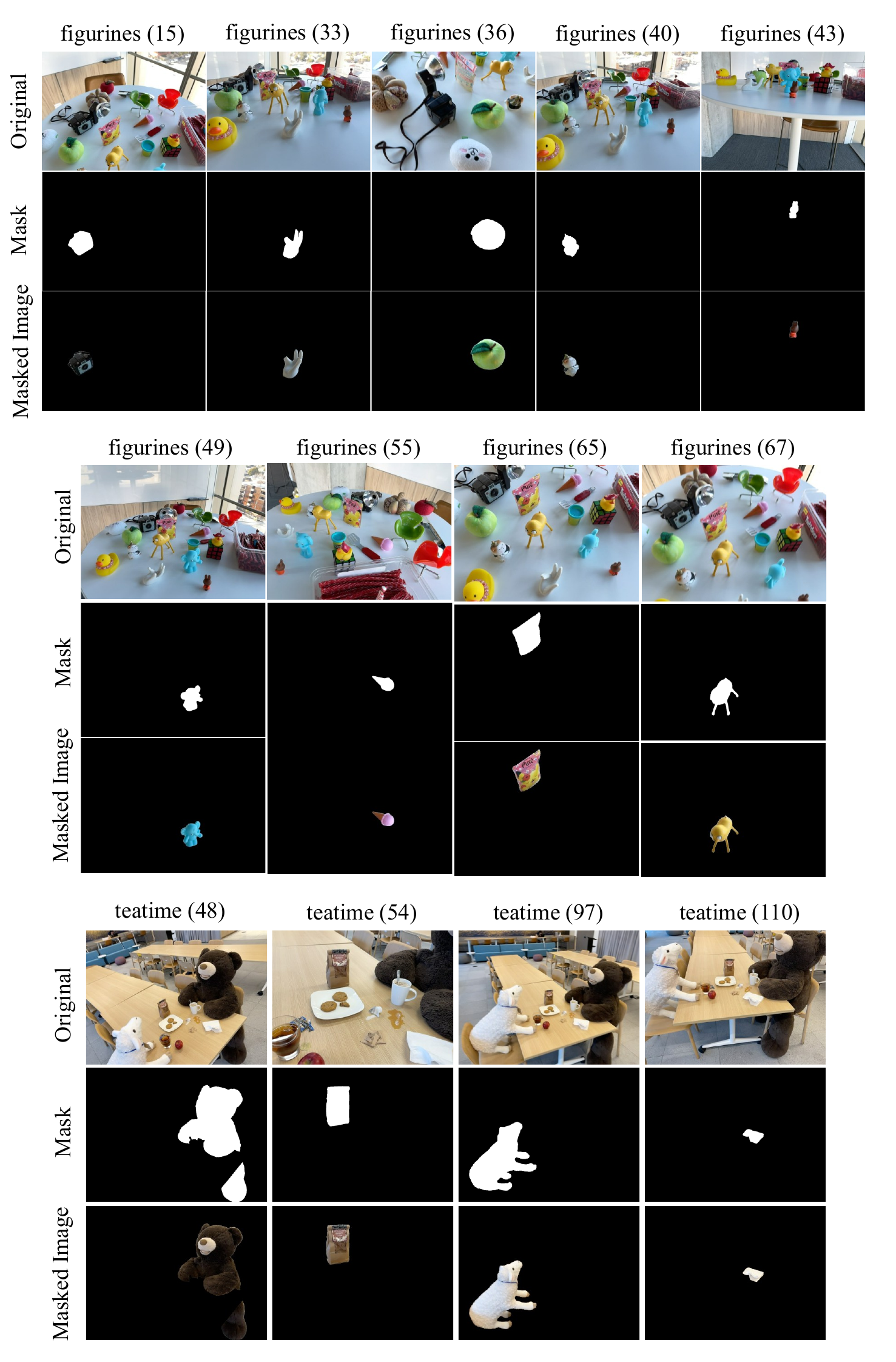}
    \caption{Examples from the LERF-Mask dataset.}
    \label{fig:lerf_mask_dataset}
\end{figure}

For Mip-NeRF 360\cite{mipnerf360} and Tanks and Temples\cite{knapitsch2017tanks}, we generate object masks using a GPT-5\cite{openai2025gpt5}--GroundingDINO\cite{liu2024grounding}--EfficientViT-SAM\cite{zhang2024efficientvitsam} pipeline, as illustrated in \cref{fig:mask_pipeline}. 
For reproducibility, we provide the example prompts used in the mask generation process. Given an input image and example prompts, GPT-5 generates an image-specific prompt that describes the target object. Example prompts are shown in \Cref{tab:example_prompt}.
Although we use GPT-5 in our implementation, this step can be replaced by other language models, since the language model is used only to generate an object description prompt.
This prompt is then used to guide GroundingDINO, which localizes the target object and produces its bounding box. The resulting bounding box is provided as the input prompt to EfficientViT-SAM to obtain the final object mask. 
To further support reproducibility, we will release the generated object prompts and mask-generation scripts.
Our method uses this pipeline only for mask generation. The proposed reconstruction framework is not tied to a specific segmentation model and can be combined with alternative segmentation methods, as shown in \cref{sec:G_0}.
For the LERF-Mask\cite{Ye2024GaussianGrouping} dataset, we use the provided multi-object masks and convert them into single-object masks based on the corresponding object IDs. 
Specifically, we extract each target object from the multi-object annotation and use the resulting binary mask as the single-object mask for training and evaluation.
The scene configurations are summarized in \Cref{tab:mipnerf}--\Cref{tab:lerf_mask}, and representative examples from the dataset are shown in \cref{fig:lerf_mask_dataset}.
In this evaluation, all methods are trained with the same masked input images and evaluated within the corresponding object regions.

\begin{table}[!t]
\centering
\caption{Example prompts provided to GPT-5 for object description generation. GPT-5 receives these example prompts together with an input image and generates a new object description in a similar style.}
\label{tab:example_prompt}
\setlength{\tabcolsep}{3pt}
\resizebox{0.7\linewidth}{!}{%
\begin{tabular}{l}
\toprule
\textbf{Example Prompt (Mip-NeRF 360)} \\
\midrule
A bicycle parked outdoors on the ground. \\
A road bike standing alone in an outdoor scene. \\
A two-wheeled bicycle with handlebars and a seat. \\
A bicycle with thin wheels leaning or standing in a park. \\
A classic bicycle frame with wheels visible in the scene. \\
A bike parked on a dirt or gravel path surrounded by trees. \\
A standalone bicycle in the center of an outdoor environment. \\
A bicycle with a metallic frame photographed from various angles. \\
\bottomrule
\end{tabular}%
}
\end{table}

\begin{table*}[!t]
    \centering
    \begin{minipage}[t]{0.46\linewidth}
        \centering
        \caption{Mip-NeRF 360 configuration.}
        \resizebox{\linewidth}{!}{%
        \renewcommand{\arraystretch}{0.85}
        \begin{tabular}{l|lcr}
        \toprule
        \textbf{Scene} & \textbf{Object} & \textbf{Object Ratio (\%)} & \textbf{Resolution} \\
        \midrule
        bicycle  & bicycle    & 15.59 & 4,946$\times$3,286 \\
        bonsai   & bonsai     & 11.39 & 3,113$\times$2,074 \\
        counter  & flower pot &  6.29 & 1,558$\times$1,038 \\
        garden   & table      & 16.89 & 5,185$\times$3,359 \\
        \bottomrule
        \end{tabular}%
        }
        \label{tab:mipnerf}
    \end{minipage}
    \hfill
    \begin{minipage}[t]{0.50\linewidth}
        \centering
        \caption{Tanks\&Temples configuration.}
        \resizebox{\linewidth}{!}{%
        \renewcommand{\arraystretch}{0.85}
        \begin{tabular}{l|lcr}
        \toprule
        \textbf{Scene} & \textbf{Object} & \textbf{Object Ratio (\%)} & \textbf{Resolution} \\
        \midrule
        Barn        & barn            & 41.75 & \multirow{4}{*}{1920$\times$1080} \\
        Caterpillar & caterpillar     & 35.84 &   \\
        Ignatius    & ignatius statue & 12.97 &   \\
        Truck       & truck           & 34.83 &   \\
        \bottomrule
        \end{tabular}%
        }
        \label{tab:tnt}
    \end{minipage}
\end{table*}

\begin{table}[!t]
\centering
\caption{LERF-Mask Configuration}
\resizebox{0.8\linewidth}{!}{%
\renewcommand{\arraystretch}{0.9}
\setlength{\tabcolsep}{7pt}
\begin{tabular}{l|lcr}
\toprule
\textbf{Scene} & \textbf{Object} & \textbf{Object Ratio (\%)} & \textbf{Resolution} \\
\midrule
figurines (15) & camera & 1.67 & \multirow{9}{*}{986$\times$728} \\
figurines (33) & hand figurine & 0.33 &  \\
figurines (36) & green apple & 1.05 &  \\
figurines (40) & cat figurine & 0.30 &  \\
figurines (43) & rabbit figurine & 0.10 &  \\
figurines (49) & elephant figurine & 0.81 &  \\
figurines (55) & ice cream figurine & 0.39 &  \\
figurines (65) & jelly bean bag & 1.16 &  \\
figurines (67) & yellow figurine & 0.86 &  \\
\midrule
teatime (48) & teddy bear & 11.76 & \multirow{4}{*}{988$\times$730} \\
teatime (54) & coffee bean bag & 0.55 &  \\
teatime (97) & sheep doll & 6.56 &  \\
teatime (110) & tissue & 0.55 &  \\
\bottomrule
\end{tabular}%
}
\label{tab:lerf_mask}
\end{table}

\begin{figure}[t]
    \centering
    \includegraphics[width=1.0\linewidth]{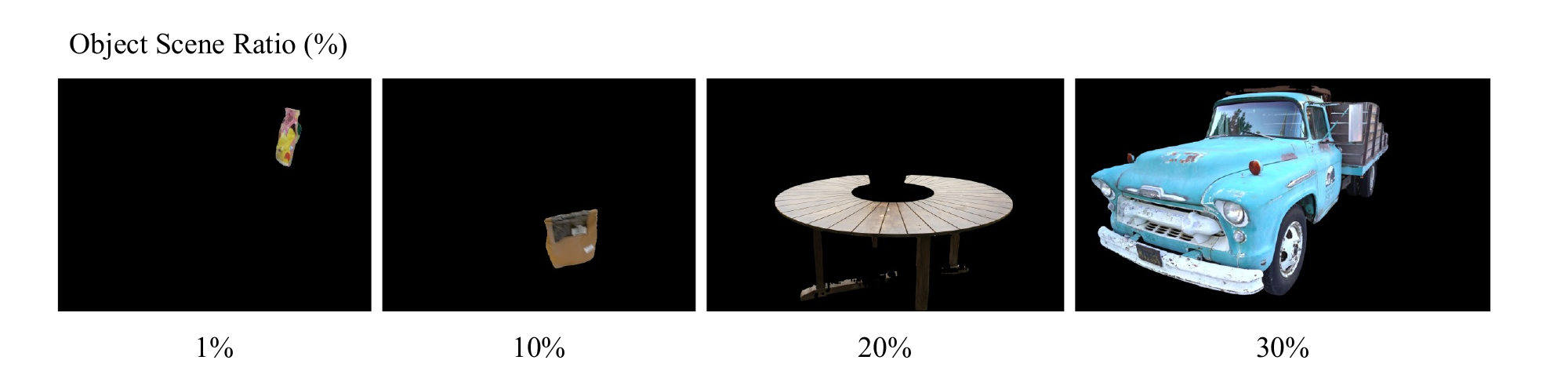}
    \caption{Visualization of object scene ratios across representative scenes. 
    Each image illustrates the proportion of the object region relative to the full scene, 
    ranging from sparse (1\%) to moderately dense (30\%).}
    \label{fig:object_ratio}
\end{figure}

\begin{figure}[t]   
    \centering
    \includegraphics[width=1.0\linewidth]{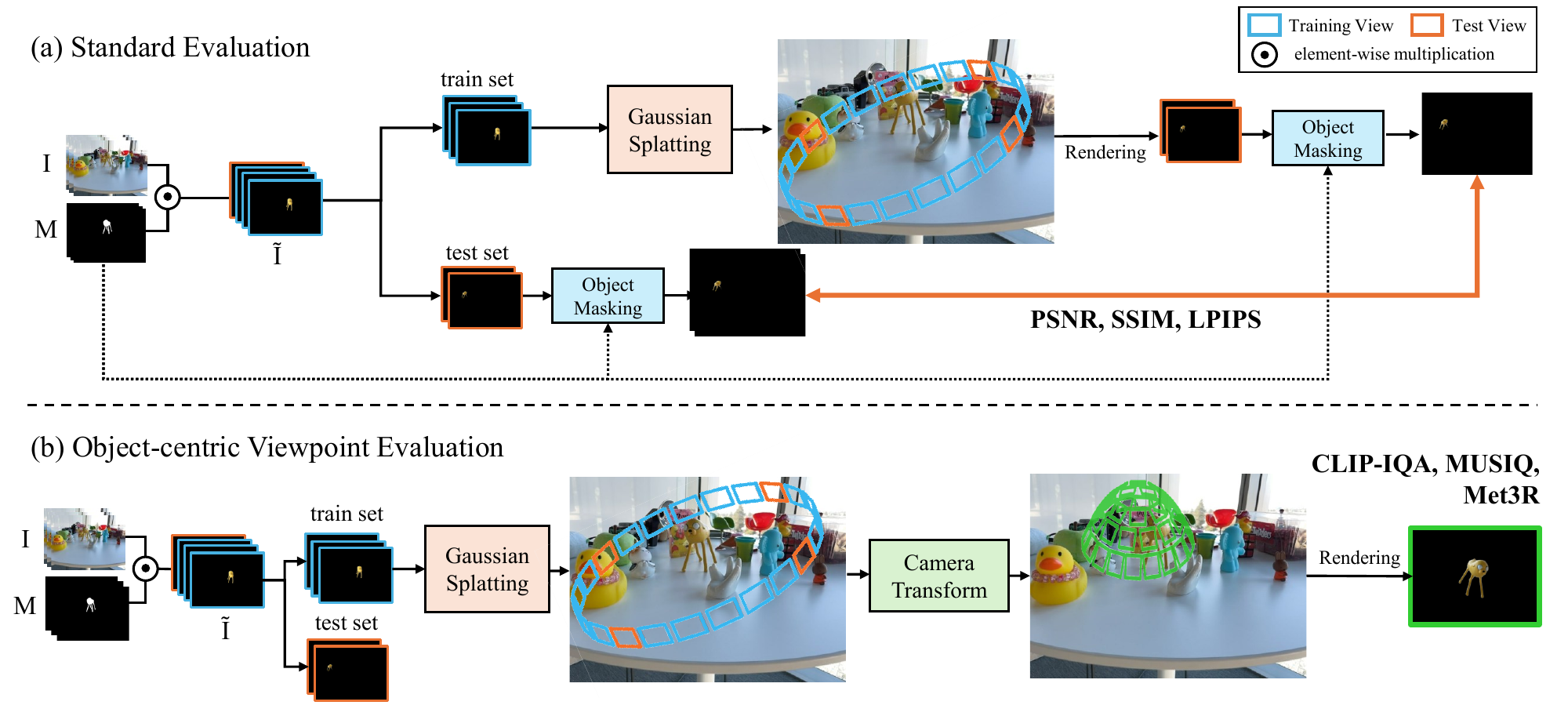}
    \caption{\textbf{Illustration of the object-centric viewpoint evaluation pipeline.} (a)~Standard Evaluation computes PSNR, SSIM, and LPIPS within the masked object regions. (b)~Object-centric viewpoint evaluation transforms camera viewpoints to zoom into the target object and assesses perceptual quality using no-reference metrics.}
    \label{fig:eval_pipeline}
\end{figure}

\subsubsection{Evaluation Metrics}
 Reconstruction quality is measured using PSNR, SSIM, and LPIPS computed exclusively within the masked object regions. As illustrated in Fig. \ref{fig:eval_pipeline}~(a), masked images $\tilde{I} = I \odot M$ are used to train the Gaussian Splatting model. At test time, object masking is applied to both the rendered output and the ground-truth images to isolate the target object region before computing the metrics. We additionally measure rendering efficiency using the FPS and the number of Gaussians.

\subsection{Evaluation of Object Boundary Quality (Sec 4.3)}
\label{sec:B_2}

\subsubsection{Datasets and Mask Annotation}
We evaluate object extraction quality on LERF-Mask~\cite{Ye2024GaussianGrouping} following the protocols of FlashSplat~\cite{shen2024flashsplat} and BEA-GS~\cite{mazzucchelli2026bea}.
LERF-Mask provides rendered-view masks for object-level supervision.
For existing object-level methods, we provide the original images containing background regions as input, following the standard evaluation setting of prior object extraction methods.
This evaluation therefore assesses post-hoc object extraction quality under the conventional full-scene input setting, rather than the masked-input reconstruction setting used for object-centric rendering evaluation.
To evaluate extraction quality, occlusion-free object silhouettes are required to capture the complete geometry of each target object.
Therefore, following the mask annotation protocol of BEA-GS, we construct additional extracted-object masks for the evaluation views.
Specifically, we use Trace3D~\cite{shen2025trace3d} outputs as visual guidance to infer the full object extent, and then manually trace and refine the final mask boundaries.
The resulting refined masks are used as the ground truth for evaluating object extraction quality.

\subsubsection{Evaluation Metrics}
This evaluation measures whether the optimized Gaussians represent the target object while excluding surrounding background regions.
We report mean IoU (mIoU) and boundary IoU (bIoU) between the extracted object mask and the ground-truth mask.
For this extraction-quality evaluation, all methods use the same object masks and train/test split, while existing object-level methods are trained with the original RGB images following the standard full-scene input setting.

\begin{table*}[!t]
\centering
\caption{DTU dataset Configuration}
\resizebox{0.9\linewidth}{!}{%
\renewcommand{\arraystretch}{0.9}
\setlength{\tabcolsep}{5pt}

\begin{tabular}{l | lccr}
\toprule
\textbf{Scene} & \textbf{Object} & \textbf{Object Ratio (\%)} & \textbf{Error Ratio (\%)} & \textbf{Resolution} \\

\midrule
scan2 & rabbit statue & 36.19 & 0.0 & 1,588$\times$1,190 \\
scan4 & bird doll & 17.86 & 0.0 & 1,587$\times$1,190 \\
scan11 & power supply & 57.27 & 6.0 & 1,583$\times$1,186 \\
scan31 & pumpkin & 22.08 & 10.0 & 1,586$\times$1,189 \\
scan32 & cacao bag & 29.61 & 6.0 & 1,587$\times$1,190 \\
scan36 & brick & 20.51 & 8.0 & 1,588$\times$1,190 \\
scan42 & cabbage & 12.56 & 6.0 & 1,587$\times$1,190 \\
scan45 & tea bag & 15.40 & 0.0 & 1,587$\times$1,189 \\
scan49 & duck doll & 37.83 & 5.0 & 1,587$\times$1,189 \\
scan54 & picture & 24.26 & 10.0 & 1,577$\times$1,176 \\
scan59 & can & 19.46 & 8.0 & 1,600$\times$1,200 \\
scan61 & chocolate milk & 17.73 & 4.0 & 1,597$\times$1,193 \\
scan75 & green pepper & 9.34 & 10.0 & 1,586$\times$1,188 \\
scan76 & canary melon & 19.71 & 0.0 & 1,585$\times$1,188 \\
scan77 & tea pot & 15.45 & 10.0 & 1,600$\times$1,200 \\
scan95 & detergent box & 24.14 & 0.0 & 1,586$\times$1,189 \\ %
scan97 & can & 9.48 & 7.0 & 1,592$\times$1,189 \\
scan103 & pig doll & 41.57 & 0.0 & 1,587$\times$1,188 \\
scan110 & bronze statue & 48.29 & 0.0 & 1,586$\times$1,189 \\
scan118 & stone statue & 31.48 & 7.6 & 1,588$\times$1,190 \\
scan123 & stone statue & 24.75 & 0.0 & 1,600$\times$1,200 \\
\midrule

scan10 & cardboard box & 46.55 & 34.0 & 1,586$\times$1,188 \\
scan12 & beverage cans & 42.75 & 46.0 & 1,588$\times$1,189 \\
scan52 & anatomical model & 41.32 & 50.0 & 1,586$\times$1,189 \\
\midrule

scan34 & bricks & 55.29 & 60.0 & 1,587$\times$1,189 \\
scan37 & scissors & 10.88 & 90.0 & 1,586$\times$1,189 \\
scan48 & mugs & 36.59 & 82.0 & 1,586$\times$1,188 \\
scan62 & volleyball & 19.44 & 80.0 & 1,585$\times$1,188 \\
\bottomrule
\end{tabular}%
}
\label{tab:dtu}
\end{table*}

\subsection{Robustness Evaluation under Mask Errors. (Sec 4.4.1)}
\label{sec:B_3}

\subsubsection{Datasets and Mask Annotation}

\begin{figure*}[t]
    \centering
    \includegraphics[width=1.0\linewidth]{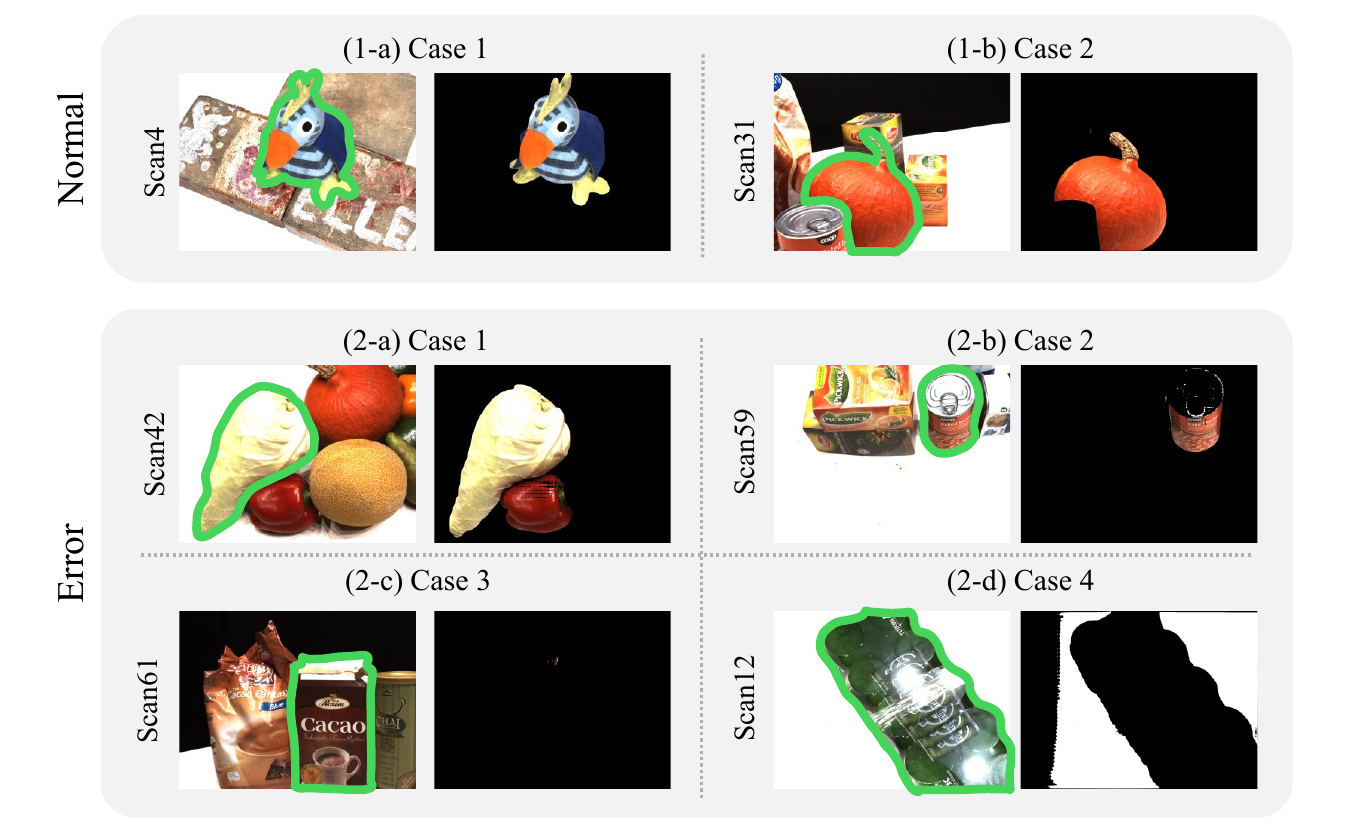}
    \caption{Representative examples of normal and erroneous mask annotations on DTU scenes. The target objects are highlighted in green.}
    \label{fig:dtu_mask_error_cases}
\end{figure*}

We use DTU~\cite{dtu} for an additional robustness analysis under mask inconsistency.
We randomly select 28 DTU scenes for this analysis, as summarized in Table~\ref{tab:dtu}, rather than choosing scenes based on specific segmentation outcomes.
For each scene, object masks are generated using EfficientViT-SAM~\cite{zhang2024efficientvitsam} from the image captured with the highest brightness level, without synthetically injecting mask errors.
Unlike the standard rendering evaluation datasets, this analysis focuses on the effect of imperfect object masks.
\cref{fig:dtu_mask_error_cases} shows representative examples of normal and erroneous mask annotations in the DTU scenes.
Normal cases include masks that accurately capture the target object while removing background regions, as shown in \cref{fig:dtu_mask_error_cases}(1-a), and masks that reasonably segment only the visible part of a partially occluded object, as shown in \cref{fig:dtu_mask_error_cases}(1-b).
We regard such occlusion-induced truncation as a normal case because the invisible part of the object cannot be reliably supervised from the corresponding view.
Erroneous cases exhibit several common types of mask inconsistency.
In \cref{fig:dtu_mask_error_cases}(2-a), the mask includes non-target regions together with the target object. In \cref{fig:dtu_mask_error_cases}(2-b), interior or boundary regions of the target object are missed, resulting in partial object truncation. In \cref{fig:dtu_mask_error_cases}(2-c), the target object is present in the image but is completely missed by the segmentation model.
In \cref{fig:dtu_mask_error_cases}(2-d), the mask includes large background regions that are unrelated to the target object. Based on these criteria, the mask error ratio is defined as the fraction of erroneous views per scene. These error patterns make DTU a suitable testbed for evaluating the robustness of object-centric reconstruction methods under realistic mask noise.

\subsubsection{Evaluation Metrics}
We group the scenes according to the mask error ratio and evaluate reconstruction quality under different levels of mask inconsistency.
The evaluation follows the same protocol as described in \cref{sec:B_1}.
The same masks are used for all methods, and no method-specific mask editing or additional supervision is applied.

\subsection{Object-centric Viewpoint Evaluation (Sec 4.4.2)}
\label{sec:B_4}

\subsubsection{Camera Trajectory Generation} %
\begin{figure}[t]
    \centering
    \includegraphics[width=0.8\linewidth]{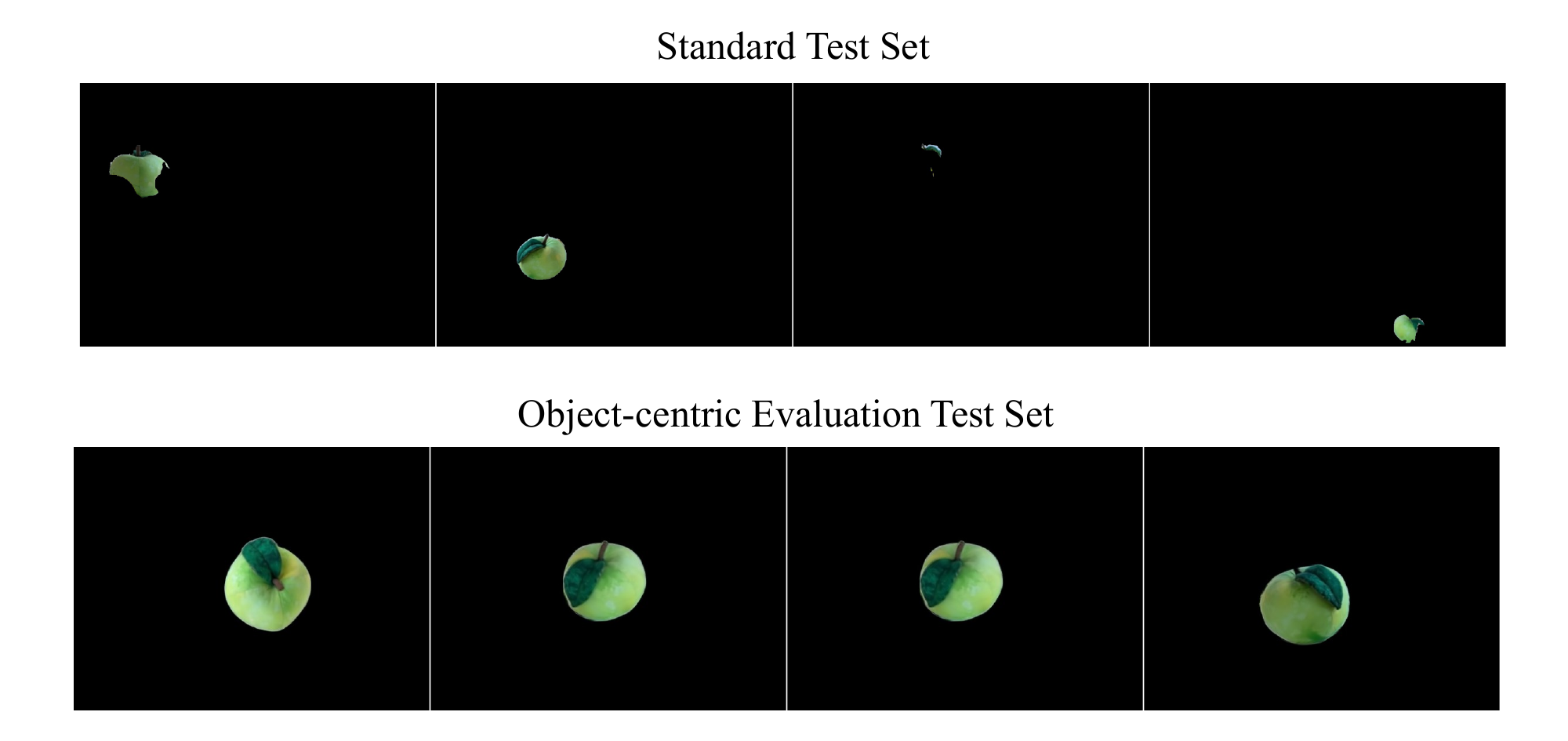}
    \caption{Visualization of the test dataset composition for the standard evaluation and the proposed object-centric evaluation.}
    \label{fig:oc_testset}
\end{figure}

As a complementary analysis, we further evaluate object-level rendering quality from object-centric viewpoints. To generate the object-centric camera poses, we construct a spherical camera trajectory centered on the target object. Given the target object instance $O$, we first compute its center $C$ and radius $R$ to determine the camera trajectory parameters. The elevation angle ranges from $\theta_{E,start}=0^\circ$ to $\theta_{E,end}=80^\circ$, uniformly sampled with $N_E=10$ elevation steps, and $N_A=20$ azimuthal views are distributed evenly over $360^\circ$. This yields a total of $N_E \times N_A = 200$ camera poses arranged in a spherical orbit around the target object. For each elevation step $j$ and azimuthal step $k$, the elevation angle $\theta_{E,j}$ and azimuthal angle $\theta_{A,k}$ are computed, and the corresponding camera-to-world transformation matrix $C2W$ is obtained via $\mathrm{ComputePose}(C,R,\theta_{E,j},\theta_{A,k})$. The resulting set of camera poses $C_{views}$ defines a dense spherical trajectory that covers the target object from multiple elevations and azimuths. This trajectory enables comprehensive evaluation of the rendering quality of the reconstructed object and its cross-view consistency.
~\cref{fig:oc_testset} compares the conventional evaluation with our object-centric evaluation protocol. Our protocol places cameras on a spherical trajectory centered on the target object and enables object-focused observations.

\subsubsection{Evaluation Metrics}
The object-centric viewpoints are rendered using the trained Gaussian representation without further optimization. Since the object-centric viewpoints do not correspond to the original camera poses in the dataset, no ground-truth images are available for the rendered views. Therefore, we evaluate the rendered results using the no-reference metrics described in the main paper. Specifically, we use CLIP-IQA+ and MUSIQ to assess image quality, and MEt3R to evaluate cross-view consistency, as illustrated in~\cref{fig:eval_pipeline}(b).
For MEt3R, we use the default configuration with MASt3R~\cite{leroy2024mast3r} as the feature extractor, DINO~\cite{caron2021emergingdino} as the backbone, FeatUp~\cite{fu2024featup} as the upsampler, and cosine distance for feature comparison. The reported score is obtained by averaging the metric values over all rendered viewpoints. Object-centric evaluation is conducted on the LERF-Mask~\cite{Ye2024GaussianGrouping} dataset using the same object instances and train/test split as the standard evaluation.

\section{Additional Experiments and Analysis}
\label{sec:C_0}

\subsection{Comparison Across Evaluation Settings}
\label{sec:eval_setting_comparison}

Table~\ref{tab:eval_setting_comparison} compares three evaluation settings on LERF-Mask covering native full-scene and object-centric reconstruction protocols.
In the \emph{Full scene} setting, ObjectGS and Trace3D are evaluated using their native full-scene pipelines.
The \emph{Extracted from full scene} setting reports object-level results obtained from these full-scene reconstructions, while the \emph{Masked input} setting follows the object-masked evaluation protocol used in Main Table~1. 

\begin{table}[t]
\centering
\caption{Comparison of reconstruction quality across evaluation settings on LERF-Mask.}
\label{tab:eval_setting_comparison}
\setlength{\tabcolsep}{7.5pt}
\begin{tabular}{l|l|ccc}
\toprule
Setting & Method & PSNR$\uparrow$ & SSIM$\uparrow$ & LPIPS$\downarrow$ \\
\midrule

\multirow{2}{*}{Full scene}
& ObjectGS~\cite{Zhu2025ObjectGS} & 23.45 & 0.860 & 0.235 \\
& Trace3D~\cite{shen2025trace3d} & 22.64 & 0.821 & 0.263 \\

\midrule

\multirow{2}{*}{\shortstack[l]{Extracted from\\full scene}}
& ObjectGS~\cite{Zhu2025ObjectGS} & 39.90 & 0.995 & 0.003 \\
& Trace3D~\cite{shen2025trace3d} & 36.36 & 0.994 & 0.005 \\

\midrule

\multirow{3}{*}{\shortstack[l]{Masked input\\(Main Table~1)}}
& ObjectGS~\cite{Zhu2025ObjectGS} & 37.65 & 0.988 & 0.013 \\
& Trace3D~\cite{shen2025trace3d} & 34.74 & 0.986 & 0.019 \\
& MaRO-GS & 42.89 & 0.992 & 0.008 \\

\bottomrule
\end{tabular}
\end{table}

\subsection{Additional Ablation Studies}
\label{sec:C_1}

While the main paper reports cumulative ablations, \Cref{tab:ablation_representative} provides representative non-cumulative combinations to further analyze the individual and interactive effects of the proposed components.
Mask-guided pruning and densification gives the most consistent single-module improvement, increasing PSNR from 41.98 to 42.55 while reducing the number of Gaussians from 42,846 to 25,805. This shows that suppressing background-supported Gaussians is essential for forming a compact object-centric representation.
Appearance-guided local Gaussian merging mainly contributes to compactness and efficiency.
Combined with pruning, it further reduces the Gaussian count to 21,157 and achieves the highest FPS with negligible PSNR degradation.
The filtering-based modules require a reliable object-centric Gaussian support to operate effectively.
Without pruning, the object region is not yet clearly localized, so geometric view consistency and region-based view validation could incorrectly filter out useful views or regions.
This over-filtering behavior explains the weak performance of the filtering-only variants, especially the large degradation caused by region-based view validation alone.
After pruning refines the object support, these filtering modules become effective: pruning with geometric view consistency improves PSNR to 42.68, and pruning with region-based view validation achieves the highest PSNR of 43.01. The full model achieves a PSNR of 42.90 with 24,422 Gaussians. Although it does not maximize a single metric, it combines reconstruction quality, compactness, efficiency, and robustness, and is therefore used as the default configuration.

\begin{table}[t]
\centering
\small
\setlength{\tabcolsep}{5pt}
\caption{
Representative ablation study of the proposed components.
The first row denotes the baseline trained with the proposed silhouette-aligned object loss only.
\textbf{P}, \textbf{M}, \textbf{G}, and \textbf{R} denote mask-guided \textbf{P}runing and densification, appearance-guided local Gaussian \textbf{M}erging, \textbf{G}eometric view consistency, and \textbf{R}egion-based view validation, respectively.
Best results are highlighted in \textbf{bold}, and the full model is \underline{underlined}.
}
\label{tab:ablation_representative}
\resizebox{0.55\linewidth}{!}{%
\begin{tabular}{cccc|ccc}
\toprule
\textbf{P}
& \textbf{M}
& \textbf{G}
& \textbf{R}
& PSNR$\uparrow$
& FPS$\uparrow$
& \#Gaussians$\downarrow$ \\
\midrule
-- & -- & -- & --
& 41.98 & 1,458 & 42,846 \\
\midrule
\checkmark & -- & -- & --
& 42.55 & 1,630 & 25,805 \\

-- & \checkmark & -- & --
& 42.38 & 1,586 & 29,966 \\

-- & -- & \checkmark & --
& 41.86 & 1,401 & 43,576 \\

-- & -- & -- & \checkmark
& 37.11 & 1,281 & 60,735 \\
\midrule

\checkmark & \checkmark & -- & --
& 42.52 & \textbf{1,784} & \textbf{21,157} \\

\checkmark & -- & \checkmark & --
& 42.68 & 1,633 & 26,951 \\

\checkmark & -- & -- & \checkmark
& \textbf{43.01} & 1,515 & 30,390 \\

\checkmark & \checkmark & \checkmark & \checkmark
& \underline{42.90} & \underline{1,653} & \underline{24,422} \\
\bottomrule
\end{tabular}%
}
\end{table}

\subsection{Analysis of Mask-Reliability View Filtering}
\label{sec:C_2}

\subsubsection{Ablation under Mask Errors}
\label{sec:mask_error_ablation}

\begin{table}[t]
\centering
\caption{View filtering ablation on the DTU splits (PSNR$\uparrow$). Geo.\ and Reg.\ denote geometric view consistency and region-based view validation, respectively.}
\label{tab:mask_error_ablation}
\setlength{\tabcolsep}{8pt}
\begin{tabular}{cc|ccc}
\toprule
Geo. & Reg. & Easy & Medium & Hard \\
\midrule
 & & 31.84 & 23.23 & 23.13 \\
\checkmark & & 31.66 & 25.68 & 23.07 \\
 & \checkmark & 33.34 & 25.98 & 23.03 \\
\checkmark & \checkmark & 33.26 & 26.50 & 22.88 \\
\bottomrule
\end{tabular}
\end{table}

To isolate the effect of the two filtering modules under mask errors, we ablate geometric view consistency and region-based view validation on the DTU Easy, Medium, and Hard splits, with all other components fixed.
\Cref{tab:mask_error_ablation} reports PSNR for each split.
Filtering is most effective on the Medium split, where corrupted and reliable views coexist, improving PSNR from 23.23 to 26.50 dB.
On the Easy split, the full filtering configuration improves PSNR from 31.84 to 33.26 dB.
On the Hard split, where most views are corrupted, filtering does not improve PSNR because the conservative criterion cannot remove enough erroneous views without sacrificing supervision.
A corresponding failure case is shown in \cref{fig:dtu_hard_failure}.

\subsubsection{Filtering Statistics}
\label{sec:filtering_statistics}

\begin{table}[t]
\centering
\caption{View-filtering statistics on the DTU splits.}
\label{tab:view_filtering_statistics}
\setlength{\tabcolsep}{5.5pt}
\resizebox{\linewidth}{!}{%
\begin{tabular}{l|c|ccc|cc}
\toprule
Difficulty &
\# Scenes &
\shortstack{Avg. \# of\\input views} &
\shortstack{Avg. \# of\\error views} &
\shortstack{Avg. \# of\\filtered views} &
Recall (\%) &
Precision (\%) \\
\midrule
Easy   & 21 & 53.3 & 2.33 & 0.8 & 24.5 & 75.0 \\
Medium & 3  & 49.0 & 21.7 & 5.0 & 23.1 & 100.0 \\
Hard   & 4  & 49.0 & 38.0 & 1.8 & 4.6 & 100.0 \\
\bottomrule
\end{tabular}
}
\end{table}

We evaluate the behavior of mask-reliability view filtering on the DTU Easy, Medium, and Hard splits.
Erroneous views are manually labeled according to the failure criteria in \cref{fig:dtu_mask_error_cases}, and these labels are used as ground truth for computing precision and recall.
Table~\ref{tab:view_filtering_statistics} reports the average numbers of input, erroneous, and filtered views for each split, together with the corresponding precision and recall.
The results show high precision with relatively low recall.
This behavior reflects a conservative filtering strategy that removes only a limited subset of unreliable views.
In particular, recall decreases to 4.6\% on the Hard split, where filtering becomes less effective as erroneous views dominate the input.
At the same time, the conservative strategy limits the irreversible removal of valid supervision.

\subsubsection{Sparse-view Analysis}
\label{sec:sparse_view_analysis}

\begin{table}[t]
\centering
\caption{Sparse-view reconstruction on DTU Scan 12 (Medium) using six input views.
Geo. and Reg. denote geometric view consistency and region-based view validation, respectively.}
\label{tab:sparse_view}
\setlength{\tabcolsep}{8pt}
\resizebox{0.6\linewidth}{!}{%
\begin{tabular}{cc|ccc}
\toprule
Geo. & Reg. & PSNR$\uparrow$ & SSIM$\uparrow$ & LPIPS$\downarrow$ \\
\midrule
     &     & 18.82 & 0.809 & 0.192 \\
\checkmark &     & 20.84 & 0.846 & 0.173 \\
     & \checkmark & 18.89 & 0.810 & 0.193 \\
\checkmark & \checkmark & 20.71 & 0.846 & 0.174 \\
\bottomrule
\end{tabular}
}
\end{table}

To examine whether mask-reliability view filtering degrades reconstruction when only a small number of input views is available, we evaluate DTU Scan 12 (Medium) using only six views.
Table~\ref{tab:sparse_view} compares geometric view consistency and region-based view validation under this sparse-view setting.
Even with only six views, applying both filtering modules does not degrade reconstruction and improves PSNR from 18.82 to 20.71 dB.

\subsubsection{Sensitivity to Filtering Schedules}
\label{sec:filtering_schedule_sensitivity}

\begin{table}[t]
\centering
\caption{Sensitivity to filtering schedules.
Geo. and Reg. denote geometric view consistency and region-based view validation, respectively.}
\label{tab:filtering_schedule}
\setlength{\tabcolsep}{4.5pt}
\resizebox{\linewidth}{!}{
\begin{tabular}{ll|ccc|ccc}
\toprule
\multicolumn{2}{c|}{Filtering Schedule} &
\multicolumn{3}{c|}{LERF-Mask} &
\multicolumn{3}{c}{DTU (Medium)} \\
\cmidrule(lr){1-2}
\cmidrule(lr){3-5}
\cmidrule(lr){6-8}
Geo. Iter. & Reg. Iter. &
PSNR$\uparrow$ & SSIM$\uparrow$ & LPIPS$\downarrow$ &
PSNR$\uparrow$ & SSIM$\uparrow$ & LPIPS$\downarrow$ \\
\midrule
1K/3K/5K (default) & 1.8K & 42.98 & 0.992 & 0.008 & 26.50 & 0.939 & 0.085 \\
1K/3K/5K           & 5.8K & 42.86 & 0.992 & 0.008 & 26.57 & 0.939 & 0.084 \\
5K/7K/9K           & 1.8K & 42.89 & 0.992 & 0.008 & 25.95 & 0.934 & 0.089 \\
5K/7K/9K           & 5.8K & 42.87 & 0.992 & 0.009 & 26.58 & 0.939 & 0.085 \\
\bottomrule
\end{tabular}
}
\end{table}

We examine whether mask-reliability view filtering depends on a fixed ordering of its two filtering modules.
We vary their schedules while keeping density refinement fixed.
Table~\ref{tab:filtering_schedule} includes a configuration where region-based view validation is applied before geometric view consistency.
Across the tested schedules, PSNR varies by only 0.12 dB on LERF-Mask and 0.63 dB on DTU Medium.
These results show that reconstruction quality is relatively insensitive to the tested filtering schedules.

\subsubsection{Sensitivity to Hit-Ratio Threshold $\tau_{\mathrm{hit}}$}
\label{sec:hit_ratio_sensitivity}
\begin{figure}[t]
    \centering
    \includegraphics[width=0.95\linewidth]{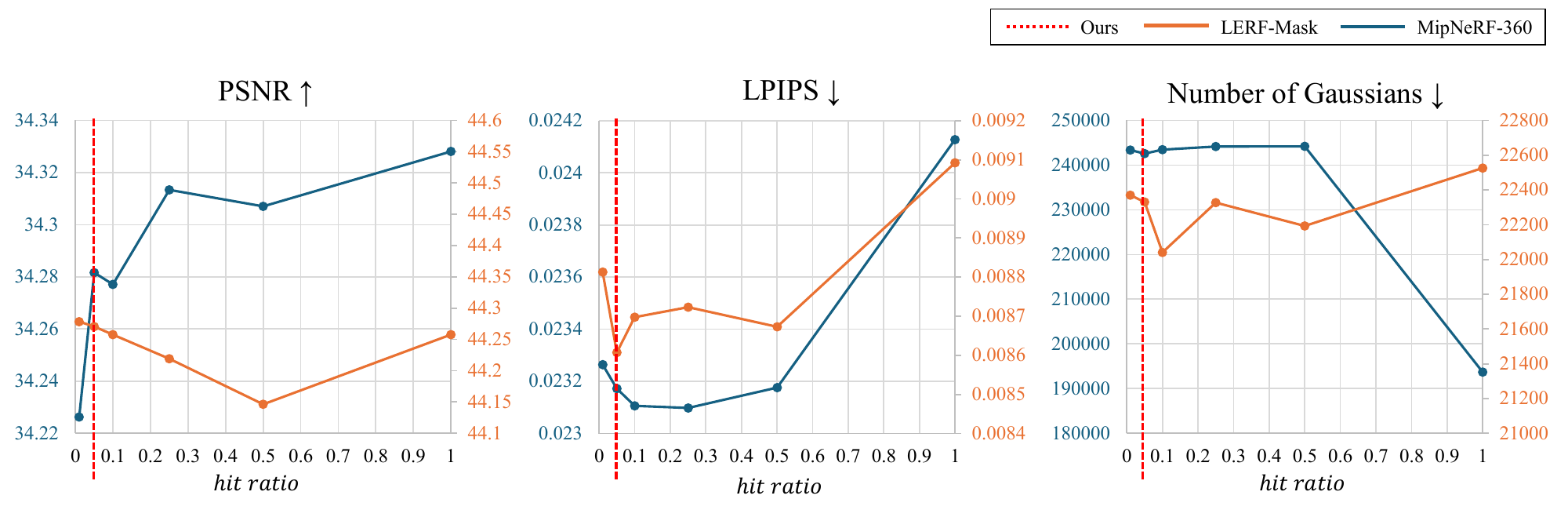}
    \caption{Sensitivity analysis of the hit-ratio threshold $\tau_{\mathrm{hit}}$. We report PSNR~$\uparrow$, LPIPS~$\downarrow$, and the number of Gaussians~$\downarrow$ across varying $\tau_{\mathrm{hit}}$ values on LERF-Mask and Mip-NeRF 360 scenes. 
    The dashed red line indicates our selected value ($\tau_{\mathrm{hit}} = 0.05$).}
    \label{fig:hitratio}
\end{figure}

We analyze the effect of the hit-ratio threshold $\tau_{\mathrm{hit}}$ on reconstruction quality and scene compactness, as shown in Fig.~\ref{fig:hitratio}. This ablation is conducted on the Mip-NeRF 360 scenes and the LERF-Mask dataset.
When $\tau_{\mathrm{hit}}$ exceeds 0.5, LPIPS increases noticeably on both datasets, indicating that overly strict view filtering removes views that still provide useful supervision. In the range $\tau_{\mathrm{hit}} \in [0, 0.5]$, PSNR remains relatively stable across both datasets.
We also observe opposite trends between the two datasets within the interval $\tau_{\mathrm{hit}} \in [0.1, 0.5]$. As $\tau_{\mathrm{hit}}$ increases, the performance on Mip-NeRF 360 slightly improves, whereas the performance on LERF-Mask decreases. This observation suggests that increasing the threshold can benefit one dataset while degrading the other.
Based on this analysis, we select $\tau_{\mathrm{hit}} = 0.05$ as the default threshold. This conservative setting avoids excessive view filtering while maintaining stable performance and compact scene representations across both datasets. All evaluations in this section were conducted on an NVIDIA RTX 4070 Ti Super GPU.

\subsection{Object-supported Gaussian Density Control Analysis}
\label{sec:C_3}

\begin{figure}[t]
    \centering
    \includegraphics[width=0.95\linewidth]{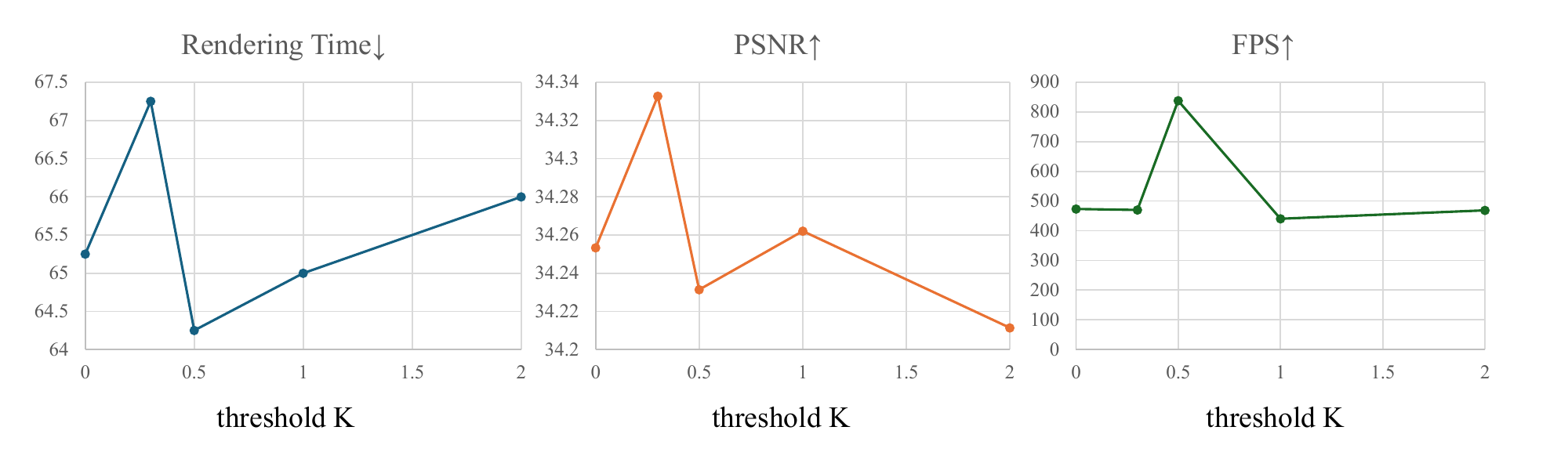}
    \caption{Ablation study on the threshold parameter $k$ used in the adaptive pruning threshold $\tau_{\text{base}}=\mu+k\sigma$ on the LERF-Mask dataset. The setting $k=0.5$ achieves the lowest rendering time and the highest FPS while maintaining comparable PSNR, and is therefore selected as the default value.}
    \label{fig:ablation_k}
\end{figure}

\begin{table}[t]
\centering
\small
\setlength{\tabcolsep}{3pt}
\caption{Sensitivity analysis of the coverage weight coefficient $\beta$ and the decay coefficient $\lambda_d$ on LERF-Mask and Mip-NeRF~360. Both coefficients show stable performance across a wide range of values (PSNR varies within 0.12 dB on LERF-Mask and 0.27 dB on Mip-NeRF 360), and are fixed as $\beta=0.5$ and $\lambda_d=0.6$ in all experiments.}
\label{tab:sensitivity}
\resizebox{0.5\linewidth}{!}{%
\begin{tabular}{c|c|cc|cc}
\toprule
\multirow{2}{*}{Parameter} & \multirow{2}{*}{Value}
  & \multicolumn{2}{c|}{LERF-Mask}
  & \multicolumn{2}{c}{Mip-NeRF 360} \\
& & PSNR$\uparrow$ & LPIPS$\downarrow$
  & PSNR$\uparrow$ & LPIPS$\downarrow$ \\
\midrule
\multirow{5}{*}{$\beta$}
& 0.00 & 44.12 & 0.0085 & 34.58 & 0.0243 \\
& 0.25 & 44.18 & 0.0085 & 34.61 & 0.0244 \\
& \textbf{0.50} & \textbf{44.17} & 0.0082 & 34.45 & 0.0244 \\
& 0.75 & 44.24 & 0.0085 & 34.54 & 0.0244 \\
& 1.00 & 44.13 & 0.0086 & 34.50 & 0.0244 \\
\midrule
\multirow{6}{*}{$\lambda_d$}
& 0.00 & 44.17 & 0.0086 & 34.57 & 0.0244 \\
& 0.20 & 44.18 & 0.0085 & 34.40 & 0.0250 \\
& 0.40 & 44.16 & 0.0086 & 34.34 & 0.0247 \\
& \textbf{0.60} & \textbf{44.25} & 0.0086 & 34.32 & 0.0253 \\
& 0.80 & 44.18 & 0.0085 & 34.59 & 0.0243 \\
& 1.00 & 44.18 & 0.0085 & 34.59 & 0.0244 \\
\bottomrule
\end{tabular}%
}
\end{table}

We analyze the behavior of the proposed object-supported Gaussian density control with respect to the adaptive pruning threshold and additional refinement hyperparameters.

\noindent\textbf{Effect of the threshold parameter $\textbf{k}$.}
\label{sec:ablation_pruning}
The parameter $k$ controls the base pruning threshold $\tau_{\text{base}}=\mu+k\sigma$, which determines how aggressively Gaussians with low mask overlap are removed. We conduct an ablation study to analyze the impact of $k$ on reconstruction quality and rendering efficiency. As shown in Fig.~\ref{fig:ablation_k}, the PSNR remains largely stable across different values of $k$, while rendering efficiency varies noticeably. In particular, $k=0.5$ achieves the lowest rendering time and the highest FPS while maintaining comparable reconstruction quality. Based on this observation, we choose $k=0.5$ as the default value since it provides the best balance between reconstruction fidelity and rendering efficiency.
Beyond the adaptive threshold parameter, we further analyze the sensitivity of additional refinement hyperparameters to evaluate the robustness of MaRO-GS under different object-centric supervision settings.

\noindent\textbf{Sensitivity to mask weight and soft decay.}
We analyze the sensitivity of the proposed refinement strategy with respect to the coverage weight coefficient $\beta$ and the decay coefficient $\lambda_d$.
The analysis is conducted on both LERF-Mask and Mip-NeRF~360 to evaluate robustness across datasets with different object area ratios. \Cref{tab:sensitivity} summarizes the results.
For $\beta$ in $w = 1 - \beta(1-\alpha)$, PSNR varies by at most 0.16\,dB across all tested values on both datasets, indicating that $\beta$ has limited influence on reconstruction quality.
We set $\beta = 0.5$ as a representative mid-range value.
For $\lambda_d$ in $\tau_{\mathrm{base}}\cdot(1-\lambda_d\,\delta\,w)$, PSNR varies by at most 0.09\,dB on LERF-Mask and 0.27\,dB on Mip-NeRF 360 across the tested values.
Since $\lambda_d$ and $\beta$ interact multiplicatively within $\tau_{\mathrm{soft}}$, their individual effects partially compensate each other, which explains the low sensitivity to either coefficient in isolation.
We set $\lambda_d = 0.6$, which achieves the best PSNR on LERF-Mask.
These results confirm that the threshold formulation does not require per-scene tuning of these coefficients.
All evaluations in this section were conducted on an NVIDIA RTX 3090.

\begin{figure*}[!t]
        \centering 
        \includegraphics[width=1.0\linewidth]{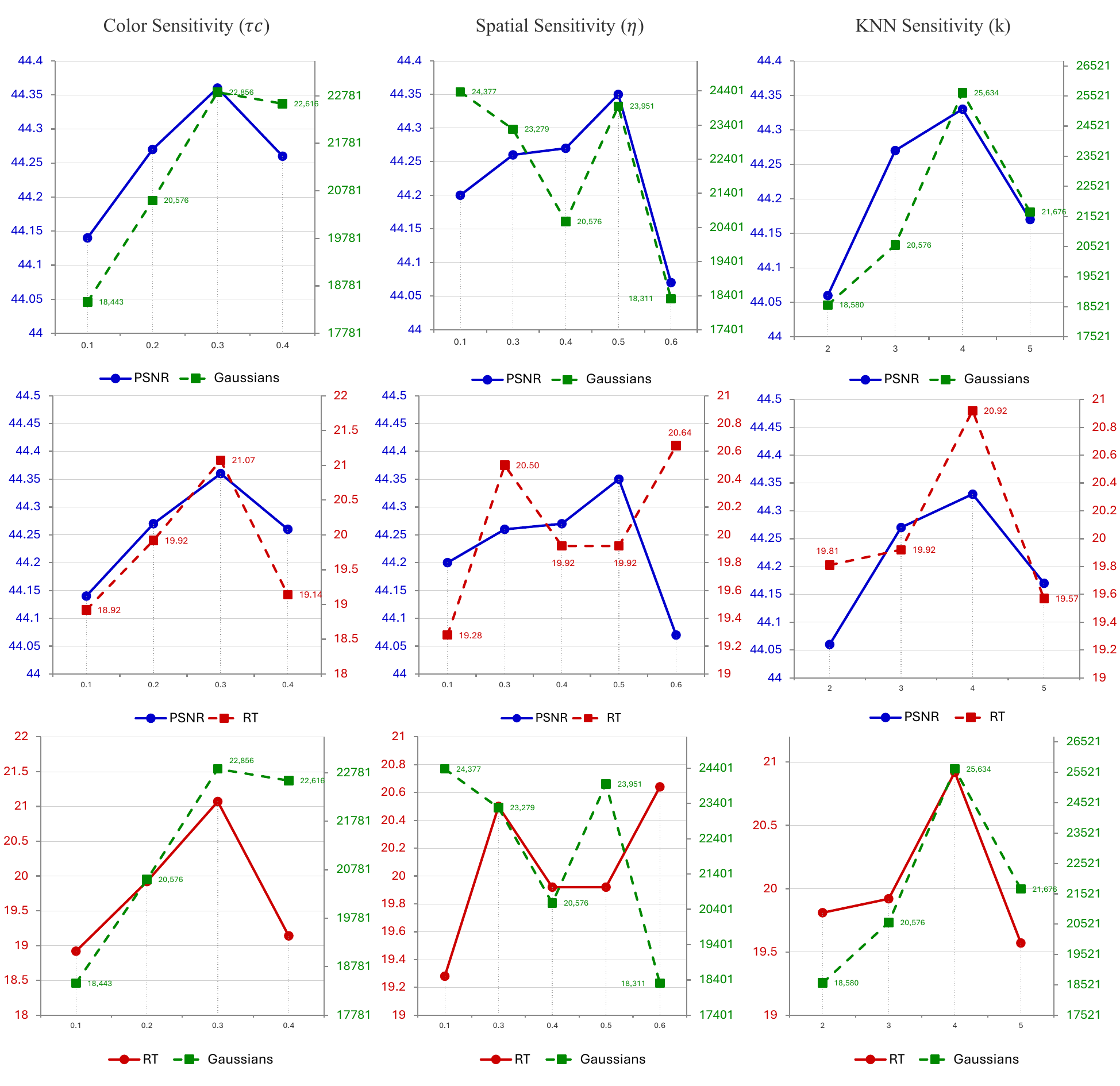}
        \caption{{\textbf{Ablation study on merging hyperparameters.} Effect of color sensitivity ($\tau_c$), spatial sensitivity ($\eta$), and group size threshold ($\kappa_{\mathrm{merge}}$) on PSNR, the number of Gaussians, and rendering time (RT). All non-varied parameters are fixed to the baseline configuration $(\tau_c, \eta, \kappa_{\mathrm{merge}}) = (0.2, 0.4, 3)$.}}
    \label{fig:merging}
\end{figure*}

\begin{figure}[!t]
    \centering
    \includegraphics[width=0.8\linewidth]{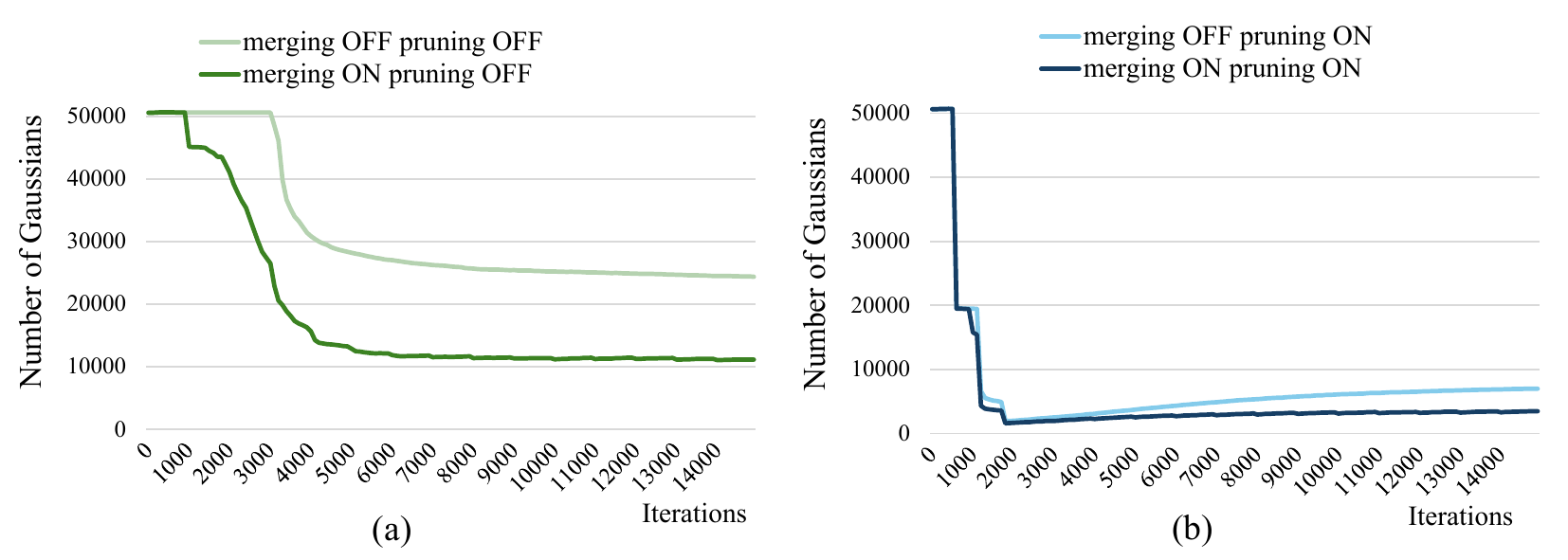} 
    \caption{Evolution of the number of Gaussians over training iterations under different configurations of appearance-guided local gaussian Merging and mask-guided pruning and densification. (a) Comparison with and without appearance-guided local gaussian Merging without mask-guided pruning and densification. (b) Comparison with and without appearance-guided local gaussian Merging when mask-guided pruning and densification is applied.}
    \label{fig:merging_pruning}
\end{figure}

\subsection{Gaussian Merging Hyperparameter Analysis}
\label{sec:C_4}
We evaluate the effect of each merging hyperparameter in Fig. \ref{fig:merging}, with a focus on rendering efficiency and compactness. 
For the color sensitivity $\tau_c$, $\tau_c=0.1$ produces the most compact representation but noticeably degrades PSNR, whereas $\tau_c=0.3$ achieves the highest PSNR at the cost of more Gaussians and higher rendering time. 
We therefore choose $\tau_c=0.2$, which offers a balanced trade-off between reconstruction quality, compactness, and rendering efficiency.
For spatial sensitivity $\eta$, $0.4$ achieves the lowest Gaussian count among configurations that maintain stable PSNR. For group size threshold $\kappa_{\mathrm{merge}}$, which controls the minimum group size required for merging, $\kappa_{\mathrm{merge}}=2$ yields the fewest Gaussians, but PSNR improves by 0.2 dB at $\kappa_{\mathrm{merge}}=3$ with a comparable Gaussian count. We therefore select $\kappa_{\mathrm{merge}}=3$. Based on these observations, the final configuration is $(\tau_c, \eta, \kappa_{\mathrm{merge}}) = (0.2, 0.4, 3)$. All evaluations in this section were conducted on an NVIDIA RTX 4070 Ti Super GPU.

\subsection{Effect of Gaussian Merging and Mask-guided Pruning}
\label{sec:C_5}

We analyze the effect of Gaussian merging and mask-guided pruning during training. 
Fig.~\ref{fig:merging_pruning} illustrates the evolution of the number of Gaussians over training iterations under different configurations. Fig.~\ref{fig:merging_pruning}(a) presents the results without mask-guided pruning while Gaussian merging is enabled or disabled. With the default configuration, Gaussian merging is performed at iterations 2K and 4K.
When merging is enabled, the number of Gaussians gradually decreases because nearby Gaussians are merged during optimization.
Fig.~\ref{fig:merging_pruning}(b) shows the results when mask-guided pruning is applied. Pruning is performed at iterations 1,000, 3,000, and 5,000 to remove redundant Gaussians. Pruning significantly reduces the number of Gaussians in the early training stage. 
When Gaussian merging and mask-guided pruning are applied together, the model achieves the most compact representation with the smallest number of Gaussians. These results demonstrate that the two modules complement each other and enable effective Gaussian compression during training.

\subsection{Silhouette-aligned Object Loss}

\begin{figure}[t]
    \centering
    \includegraphics[width=1.0\linewidth]{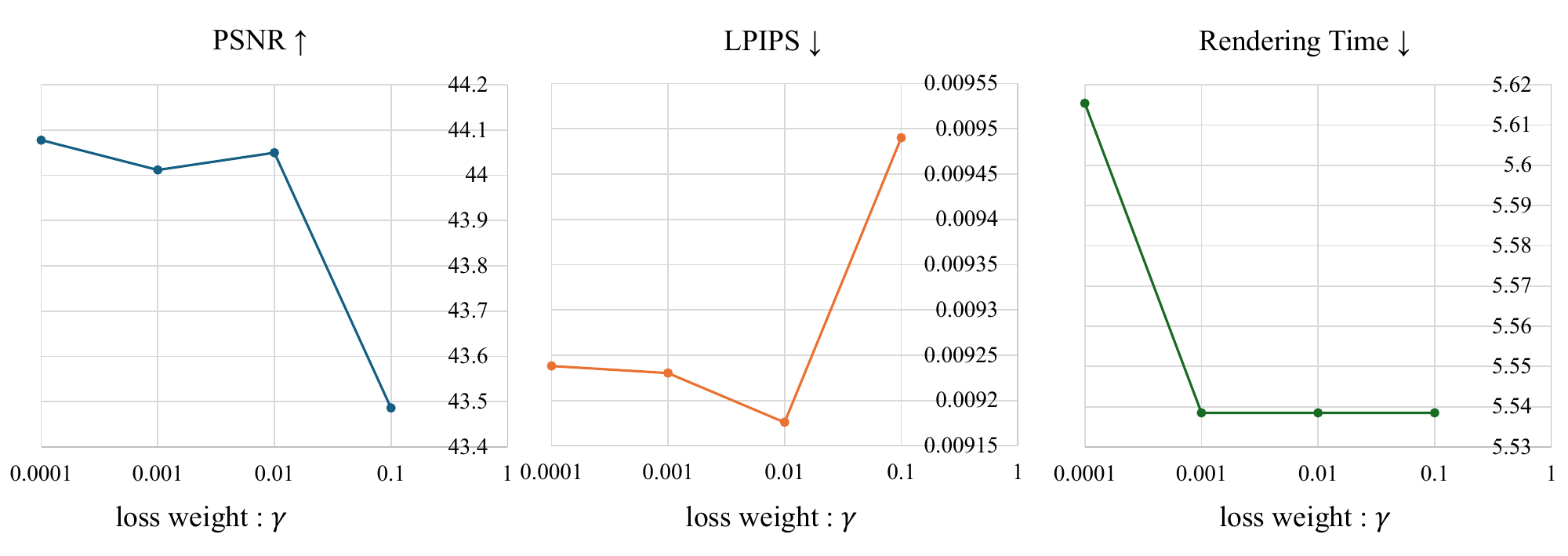}
    \caption{Ablation study on the silhouette-aligned object loss weight $\gamma$. 
    $\gamma=0.01$ provides the best trade-off between reconstruction quality and rendering efficiency.}
    \label{fig:loss}
\end{figure}

We analyze the effect of the background loss weight $\gamma$, which controls the contribution of background regions during training. 
When $\gamma=0$, the model is optimized using only the masked object region, completely ignoring the background. 
Although this enforces strong object-centric supervision, it can reduce training stability because surrounding scene context is not considered. Introducing a small background weight improves reconstruction stability while still emphasizing the target object. 
As shown in Fig. \ref{fig:loss}, $\gamma=0.01$ maintains high PSNR and low LPIPS while preserving efficient rendering time. 
In contrast, when $\gamma$ becomes too large ($\gamma=0.1$), the training objective becomes dominated by background regions, which weakens the object-centric constraint and degrades reconstruction quality.
Based on these observations, we select $\gamma=0.01$ as the default setting since it provides the best trade-off between object-focused supervision and stable scene optimization. All evaluations in this section were conducted on an NVIDIA RTX 3090 GPU.

\section{Problem Analysis: Challenges in Object-Centric 3D Gaussian Splatting}
\label{sec:D_0}

\subsection{Challenges Under Inconsistent Supervision}
\label{sec:D_1}

\begin{table}[t]
\centering
\caption{Error rate comparison across datasets}
\label{tab:errorrate}
\setlength{\tabcolsep}{7pt}
\resizebox{0.6\linewidth}{!}{%
\begin{tabular}{c|cccc}
\toprule
\multicolumn{5}{c}{\textbf{Tanks and Temples}} \\
\midrule
\textbf{Scene} 
& Barn & Caterpillar & Ignatius & Truck \\
\midrule
\textbf{Error Rate} 
& 30.97 & 62.14 & 25.70 & 52.58 \\
\midrule
\multicolumn{5}{c}{\textbf{Mip-NeRF 360}} \\
\midrule
\textbf{Scene} 
& bicycle & bonsai & counter & garden \\
\midrule
\textbf{Error Rate} 
& 80.51 & 72.26 & 52.50 & 61.08 \\
\bottomrule
\end{tabular}
}
\end{table}

As discussed in Sec.~3.1 of the main paper, object-centric Gaussian reconstruction requires accurate object region information. However, obtaining perfect segmentation masks in practice is nearly impossible. \Cref{tab:errorrate} reports the mask error rates on the Tanks and Temples and Mip-NeRF 360 datasets when masks are generated automatically with the pipeline described in \cref{fig:mask_pipeline}. The error rates range from 25\% to over 80\%. These results indicate that object regions are frequently masked incorrectly even under a standard automatic pipeline. When these noisy masks are directly used in 3DGS, the errors propagate throughout the optimization process and degrade rendering quality, as illustrated in Fig.~3 of the main paper.

\subsection{Background Removal Artifacts}
\label{sec:D_2}

\begin{figure*}[!t]
    \centering
    \includegraphics[width=0.9\linewidth]{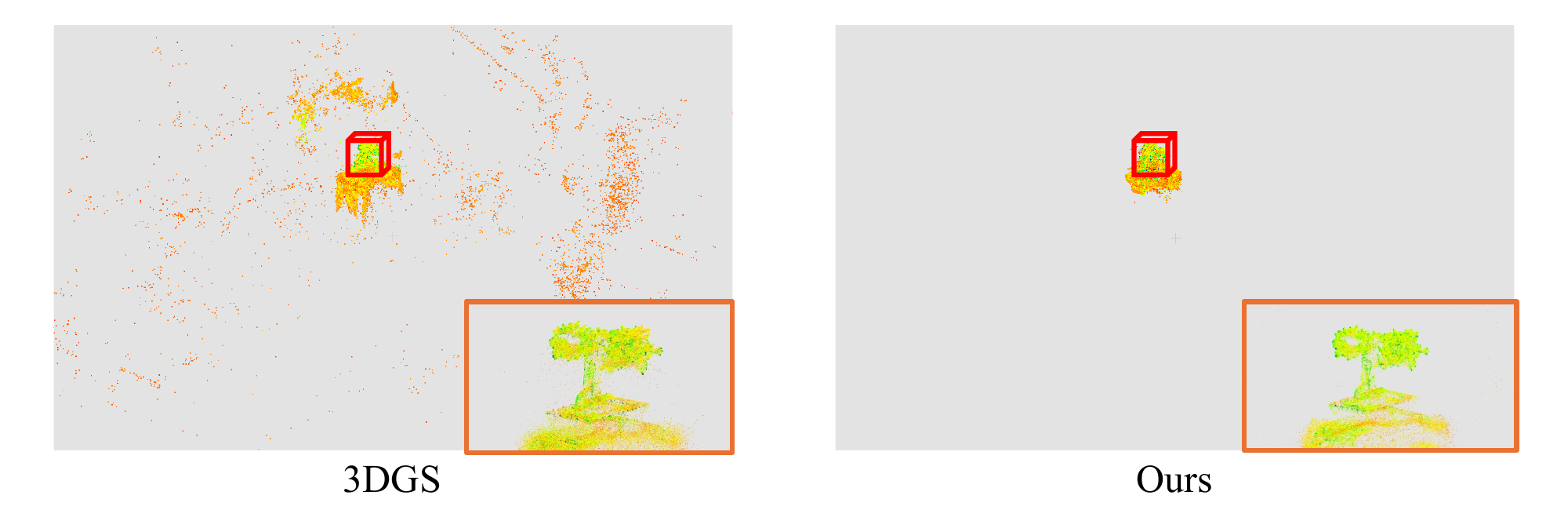}
    \caption{Visualization of Gaussian central points after background removal. The red bounding box indicates the target object region.}
    \label{fig:back_points}
\end{figure*}

When background regions are removed, conventional 3DGS still generates a large number of Gaussians in empty regions. \cref{fig:back_points} visualizes the Gaussian distributions after background masking. Although the background is removed, the baseline 3DGS reconstructs a large spatial region around the target object and produces many redundant Gaussians in the surrounding space. In contrast, our method maintains a compact Gaussian distribution that is concentrated around the target object. The presence of redundant Gaussians can also be observed in the rendered results. 
As shown in the truck in \cref{fig_quality_tanks}, objects exhibit white regions where parts of the object are partially removed or occluded. These artifacts arise from redundant Gaussians generated in surrounding empty regions. 
In contrast, our method suppresses these redundant Gaussians and produces more complete object reconstruction.

\section{Detailed Comparison with Related Work}
\label{sec:E_0}

\subsection{Comparison with ObjectGS}
\label{sec:E_1}

\begin{table*}[!t]
\centering

\caption{Object Reconstruction Strategies: ObjectGS vs. MaRO-GS.}
\label{tab:ObjectGS}
\resizebox{0.9\linewidth}{!}{%
\renewcommand{\arraystretch}{1.0}
\begin{tabular}{|l|p{6cm}|p{6cm}|}
\hline
 & \textbf{ObjectGS~\cite{Zhu2025ObjectGS}} & \textbf{MaRO-GS (Ours)} \\ \hline
\textbf{Main Task} & Object-aware scene reconstruction and understanding &  Robust object-centric reconstruction under inconsistent mask supervision \\ \hline
\textbf{Input Data} & 2D multi-view images + 2D instance masks  & Object-masked multi-view images + 2D object masks \\ \hline
\textbf{Problem Formulation} & Semantic ambiguity in scene-level rendering & Rendering instability under inconsistent mask supervision \\ \hline
\textbf{Role of Masks} & Semantic supervision for object-level classification & Object supervision for Gaussian optimization \\ \hline
\end{tabular}%
}

\vspace{0.5cm} %

\caption{Pruning-based GS Comparisons: PUP 3D-GS vs.\ MaRO-GS.}
\label{sec:E_2}

\label{tab:PUP}
\resizebox{0.9\linewidth}{!}{%
\renewcommand{\arraystretch}{1.0}
\begin{tabular}{|l|p{6cm}|p{6cm}|}
\hline
 & \textbf{PUP 3D-GS}~\cite{pup} & \textbf{MaRO-GS (Ours)} \\ \hline
\textbf{Main Task}
 & Post-hoc compression of pretrained 3DGS models for efficient storage and rendering
& Object-centric reconstruction from object-masked multi-view images \\ \hline
\textbf{Object Preservation}
 & Implicit foreground retention via multi-view sensitivity.
 & Explicit object preservation via mask-guided pruning and densification.\\ \hline
\textbf{Pruning Strategy}
  & Low-sensitivity Gaussians are pruned based on reconstruction sensitivity.
 & Unsupported Gaussians are pruned based on mask overlap. \\ \hline
\end{tabular}%
}

\vspace{0.5cm} %

\caption{Covisibility Usage: LongSplat vs.\ MaRO-GS.}
\label{tab:covisibility}
\resizebox{0.9\linewidth}{!}{%
\renewcommand{\arraystretch}{1.0}
\begin{tabular}{|l|p{6cm}|p{6cm}|}
\hline
 & \textbf{LongSplat~\cite{lin2025longsplat}} & \textbf{MaRO-GS (Ours)} \\ \hline
\textbf{Input}
& Gaussians in consecutive frames $t$ and $t'$.
& Gaussians in training view $v_i$ and its reference views. \\ \hline
\textbf{How}
& IoU of binary visible-Gaussian sets between two frames.
& Soft IoU of Gaussian visibility weighted by mask value and opacity, averaged over TopK reference views. \\ \hline
\textbf{Why}
& To decide which nearby frames should be jointly optimized to ensure reliable multi-view supervision.
& To determine which training views should be excluded to ensure reliable object-level supervision. \\ \hline

\end{tabular}%
}

\end{table*}

Table \ref{tab:ObjectGS} summarizes the differences between ObjectGS~\cite{Zhu2025ObjectGS} and MaRO-GS. ObjectGS performs scene-level reconstruction and leverages 2D instance masks as semantic supervision for object-level classification via one-hot ID encoding. In contrast, MaRO-GS reconstructs objects directly from object-masked multi-view images, where masks suppress background Gaussians and filter inconsistent views.

\subsection{Comparison with PUP 3D-GS}
\label{sec:E_3}
Table \ref{tab:PUP} compares PUP 3D-GS\cite{pup} and MaRO-GS. PUP 3D-GS focuses on compressing pretrained 3D-GS models via sensitivity-based pruning. The authors suggest that foreground Gaussians tend to be implicitly retained because they are observed across more training views. In contrast, MaRO-GS explicitly preserves the target object through mask-guided pruning for object-centric reconstruction.

\subsection{Comparison with LongSplat}
Table~\ref{tab:covisibility} summarizes the differences in covisibility usage between LongSplat~\cite{lin2025longsplat} and MaRO-GS. While both methods utilize Gaussian visibility to measure view-to-view relationships, they serve fundamentally different purposes in their respective pipelines. LongSplat computes a binary IoU of visible-Gaussian sets between frame pairs, and uses it to decide which nearby frames should be jointly optimized within a visibility-adapted local window, ensuring reliable multi-view supervision during incremental pose and scene optimization.
In contrast, MaRO-GS leverages Gaussian visibility IoU across reference views as a reliability metric for view consistency. By evaluating the overlap between a target training view and its geometrically related reference views, MaRO-GS effectively discards geometrically inconsistent views that might otherwise introduce artifacts. This mechanism ensures robust object-centric reconstruction, particularly when dealing with noisy or inconsistent mask supervision that commonly arises from automated segmentation models such as SAM.

\section{Limitations and Failure Case Analysis}
\label{sec:F_0}

\label{sec:supple_H}
\begin{figure}[t]
    \centering
    \includegraphics[width=1.0\linewidth]{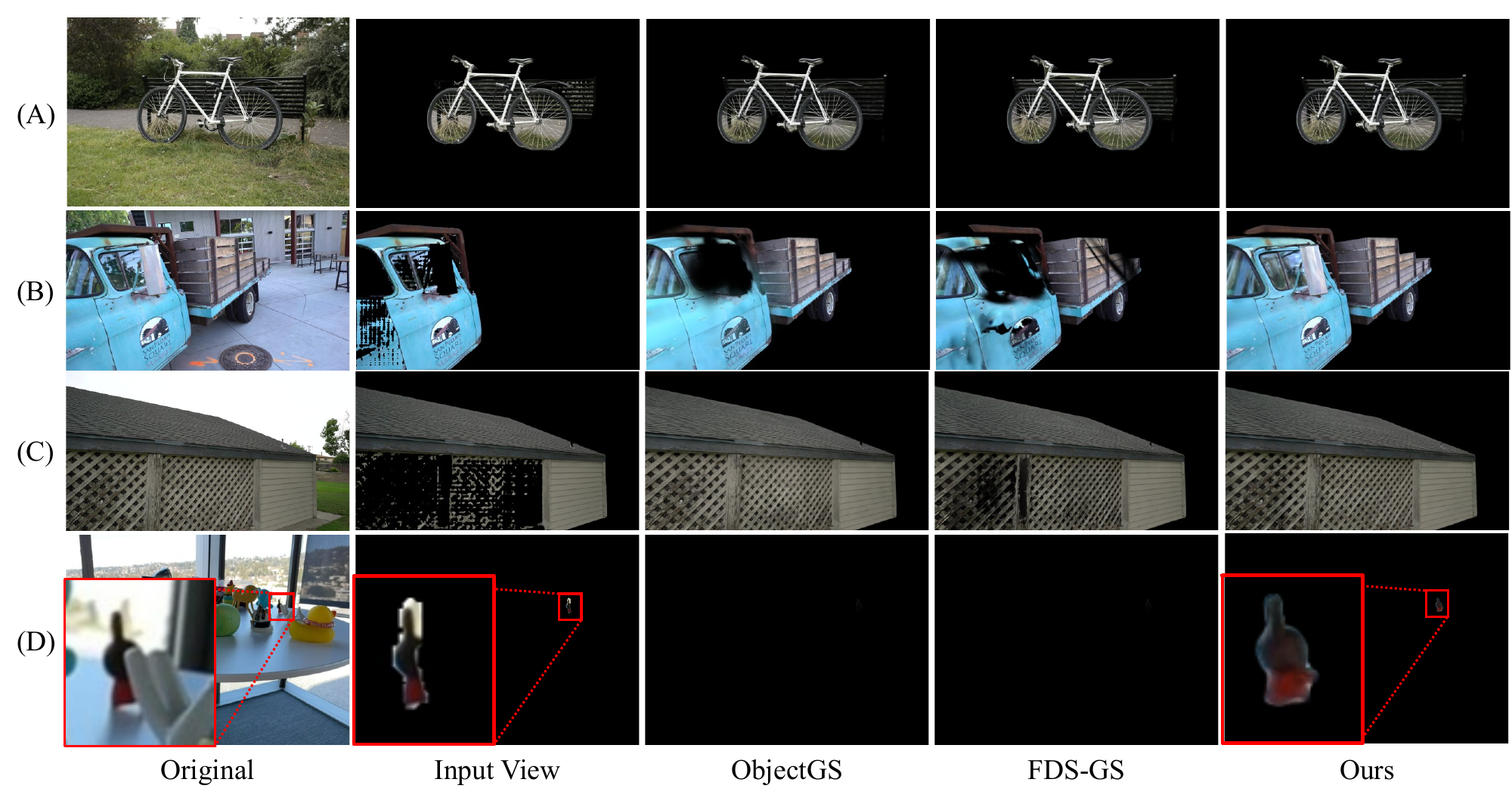}
    \caption{\textbf{Challenging cases of MaRO-GS.} (A) Overlapping objects preventing clean isolation of the target object. (B–C)  Inaccurate object masks limiting quantitative evaluation despite improved visual quality. (D) Extremely small target object, where perceptual improvements are underestimated by quantitative metrics.}
    \label{fig:limitation}
\end{figure}

\begin{figure}[!t]
    \centering
    \resizebox{0.7\linewidth}{!}{%
    \includegraphics[width=1.0\linewidth]{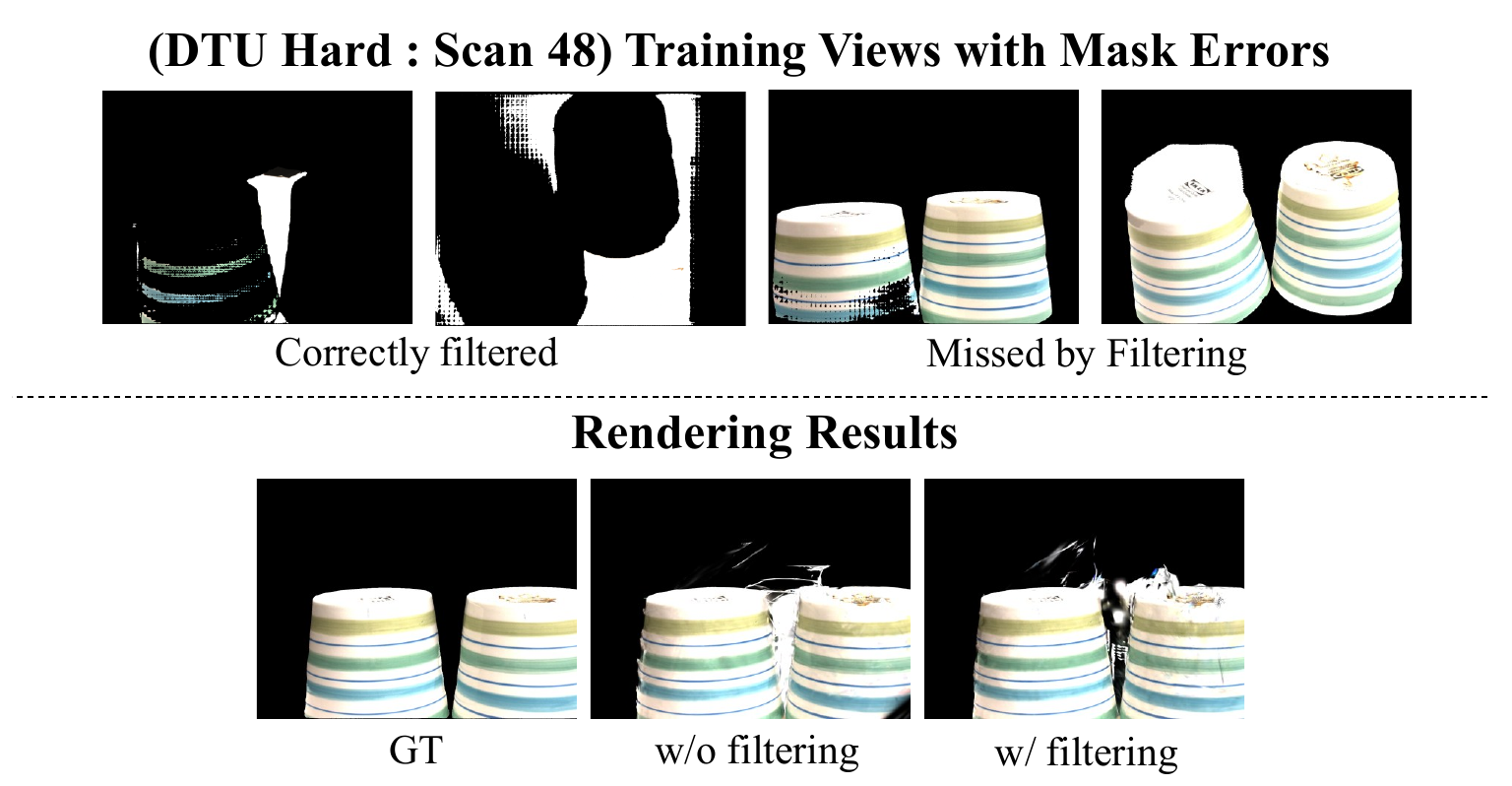}
    }
     \caption{Failure case on the DTU Hard split (Scan 48).
Top: training views with mask errors. Some views are correctly removed by filtering, while others are missed.
Bottom: rendering results without and with filtering.
Residual artifacts remain due to erroneous views that are not filtered.}
    \label{fig:dtu_hard_failure}
\end{figure}

\subsection{Challenging Cases and Limitations of MaRO-GS}
\label{sec:F_1}

Although our method achieves robust object-centric reconstruction across diverse scenes, several challenging cases remain, as illustrated in \cref{fig:limitation}--\cref{fig:dtu_hard_failure}.
In \cref{fig:limitation}(A), the target object overlaps with a nearby object, where thin structures (e.g., bicycle spokes) and overlapping regions are consistently merged by the segmentation masks across almost all viewpoints. Since our framework is designed to mitigate view-level inconsistent supervision, it cannot recover when incorrect supervision is consistently propagated across nearly the entire view set. Consequently, the optimization receives incorrect object supervision throughout training, resulting in the joint reconstruction of both objects. 
In \cref{fig:limitation}(B)--(C), the provided object masks contain inaccurate object boundaries. Because the standard evaluation computes quantitative metrics only over the object-mask regions, visual improvements beyond the inaccurate mask boundaries are not reflected in the reported scores. As a result, MaRO-GS produces perceptually more accurate reconstructions, while the quantitative metrics remain comparable to those of the baseline methods.
In \cref{fig:limitation}(D), the target object occupies only a very small image region and is visible in only a limited number of views. Baseline methods fail to reconstruct a meaningful object structure under this setting, whereas MaRO-GS still recovers a coherent object shape. Nevertheless, this substantial qualitative gap is reflected in only a marginal difference in the quantitative scores.

\cref{fig:dtu_hard_failure} further illustrates a failure case on the DTU Hard split (Scan 48).
The filtering module removes some erroneous views, but several unreliable views remain unfiltered.
These missed error views continue to provide incorrect supervision during optimization and cause residual artifacts in the reconstructed object.
This case reflects the limitation of conservative filtering under severe mask corruption, where not all erroneous views can be removed reliably.

These cases indicate that MaRO-GS still depends on the quality of the input segmentation masks under challenging conditions, while also highlighting limitations of the current evaluation protocol. Improving robustness to imperfect segmentation, together with more reliable object-centric datasets and evaluation protocols, remains an important direction for future work.

\subsection{Failure Cases of Existing Methods}
\label{sec:F_2}

\begin{figure}[t]
    \centering
    \begin{minipage}[t]{0.70\linewidth}
        \centering
        \includegraphics[width=\linewidth]{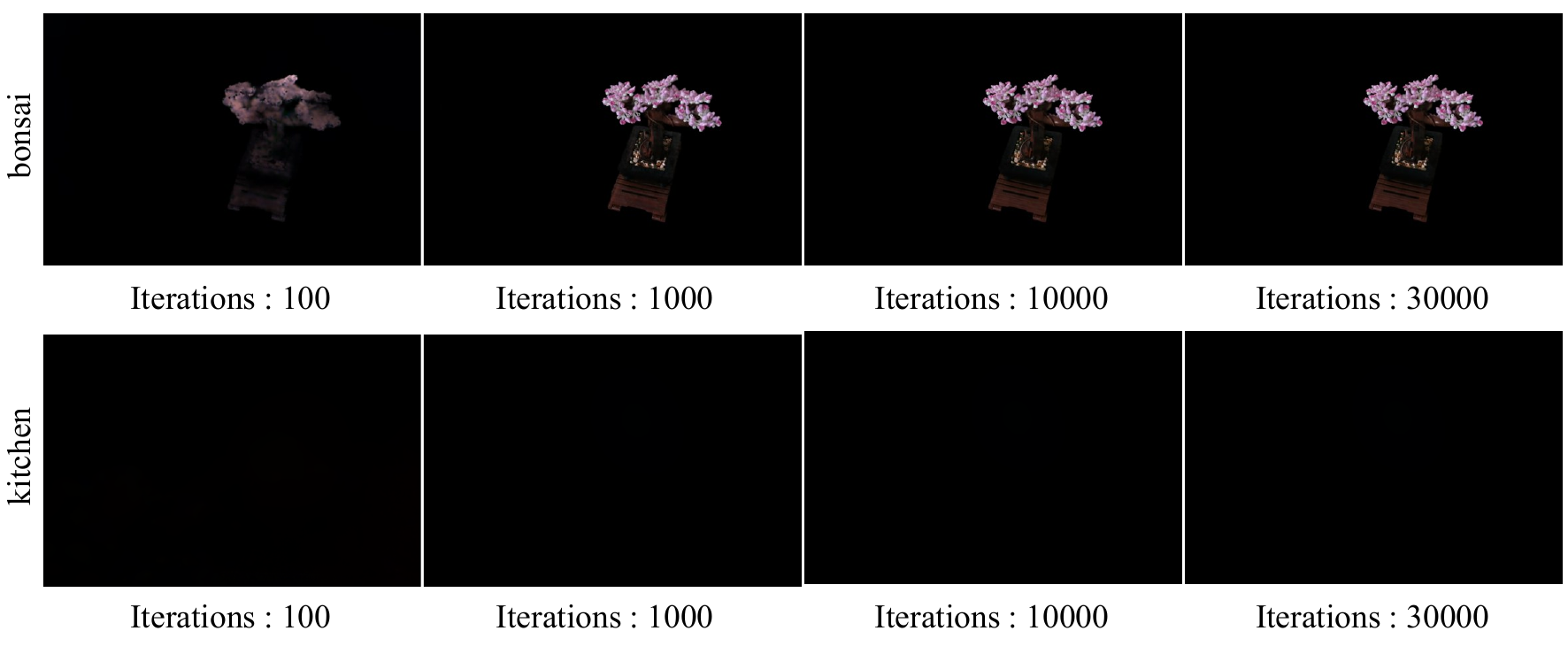}
        \caption{Failure case of ObjectGS on the kitchen scene from Mip-NeRF 360. While the bonsai scene is reconstructed successfully, the kitchen scene produces no visible object throughout training.}
        \label{fig:obj_failure}
    \end{minipage}
    \hfill
    \begin{minipage}[t]{0.28\linewidth}
        \centering
        \includegraphics[width=0.7\linewidth]{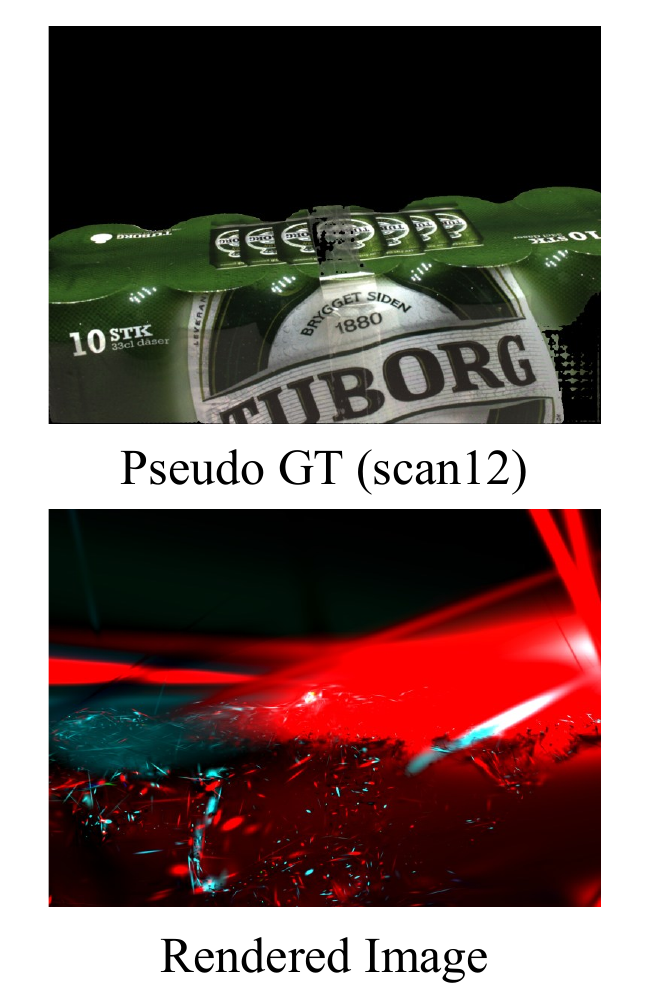}
        \caption{Failure case of PUP 3D-GS on DTU scenes.}
        \label{fig:pup_fail}
    \end{minipage}
\end{figure}

In the kitchen scene on Mip-NeRF 360 dataset, ObjectGS~\cite{Zhu2025ObjectGS} failed to reconstruct any visible object when this scene was used as input, as shown in Fig. \ref{fig:obj_failure}. 
ObjectGS is designed to optimize object-centric Gaussians based on object segmentation priors. 
When the object region is small or ambiguous, the method may fail to initialize reliable object Gaussians, which can lead to an empty reconstruction. 
PUP 3D-GS~\cite{pup} showed stable behavior on Mip-NeRF 360 and LERF-Mask but exhibited a failure case on the DTU dataset. In several DTU scenes, the method pruned most Gaussians in the scene during training. Since DTU contains many views with inconsistent geometry and background clutter, the pruning strategy can incorrectly suppress Gaussians that correspond to valid object structures. This result indicates that pruning-based acceleration methods may become unstable when the importance of scene elements cannot be reliably estimated. \cref{fig:pup_fail} shows that even in the few scenes where training successfully completes, the reconstructed results remain inaccurate.

\begin{figure}[t]
    \centering
    \resizebox{0.8\linewidth}{!}{%
    \includegraphics[width=1.0\linewidth]{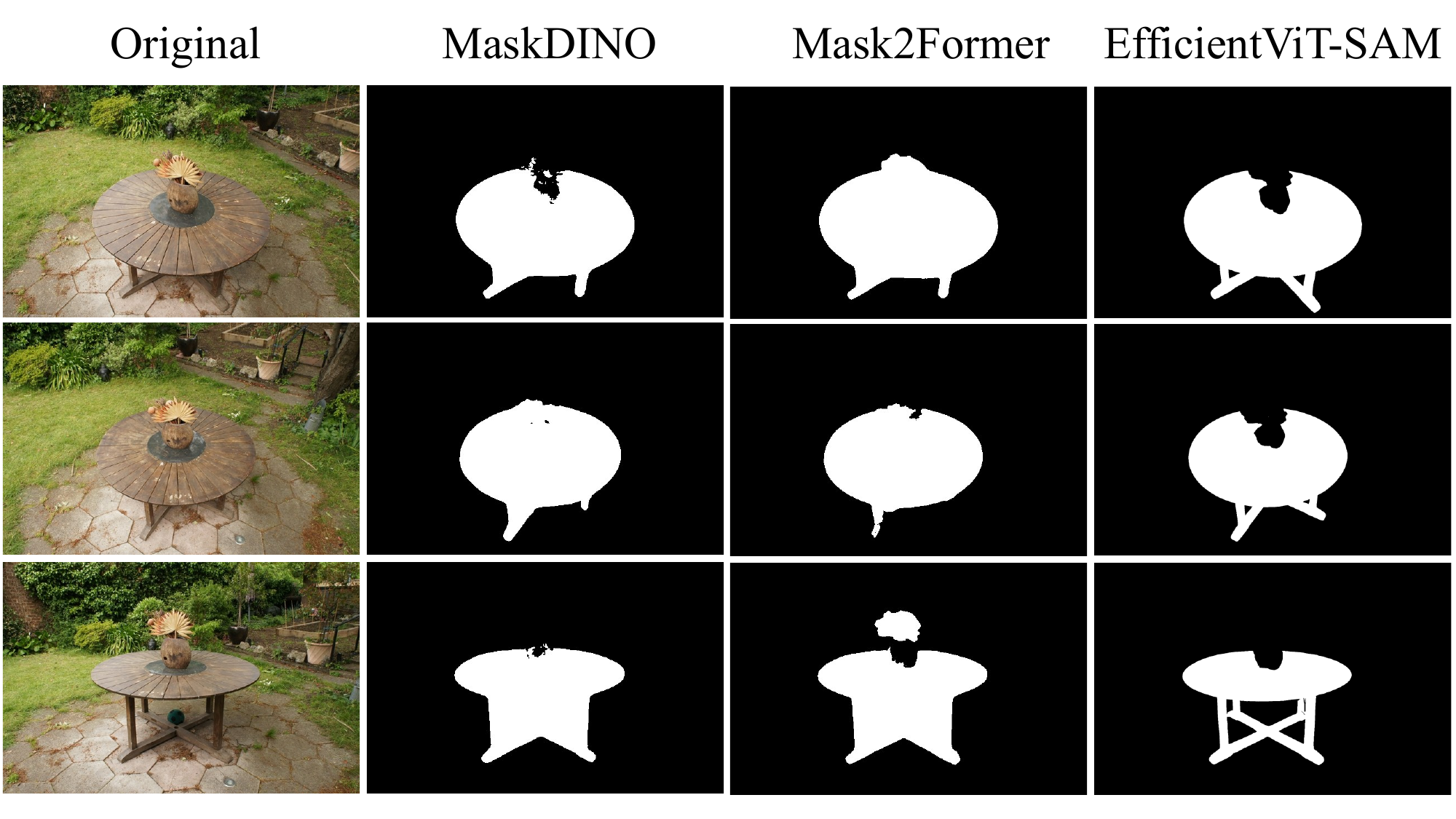}
    }
    \caption{EfficientViT-SAM yields more precise object boundaries, while Mask2Former and MaskDINO show noticeable background leakage around thin structures.}
    \label{fig:supple_seg_mask}
\end{figure}

\begin{figure}[!t]
    \centering
    \resizebox{1.0\linewidth}{!}{%
    \includegraphics[width=1.0\linewidth]{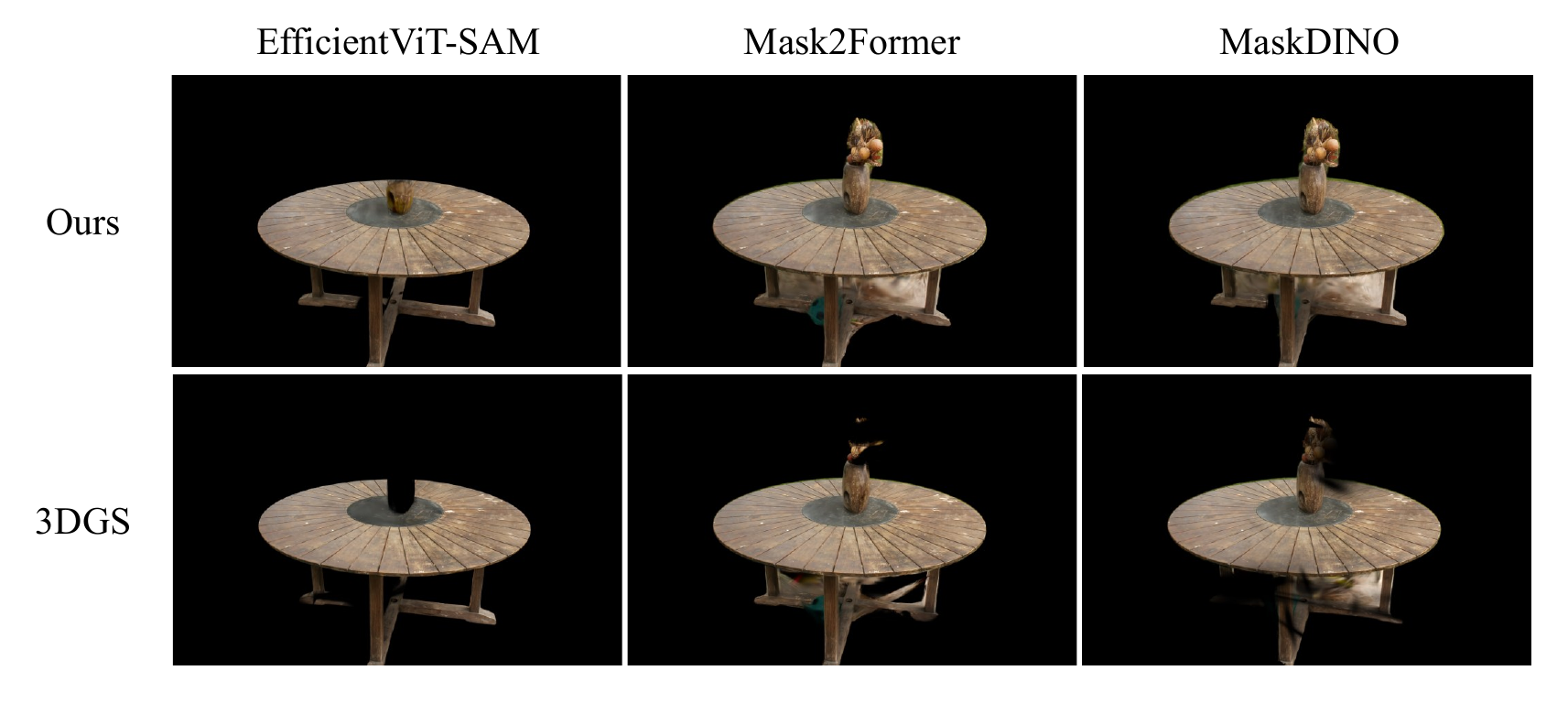}
    }
    \caption{Comparison of object masks and reconstruction results
using different segmentation models.}
    \label{fig:supple_RENDER}
    
\end{figure}

\section{Robustness Across Segmentation Models}
\label{sec:G_0}

For datasets requiring generated masks in the main paper, our framework uses EfficientViT-SAM~\cite{zhang2024efficientvitsam} as the segmentation model due to its strong segmentation quality and computational efficiency. To evaluate the robustness of MaRO-GS under different segmentation models, we additionally conduct experiments using masks generated by Mask2Former~\cite{cheng2022maske2FORMER} and MaskDINO~\cite{li2023maskdino}. As shown in Fig.~\ref{fig:supple_seg_mask}, masks generated by Mask2Former and MaskDINO exhibit less precise object boundaries, with noticeable background leakage around thin structures such as table legs.
Fig.~\ref{fig:supple_RENDER} compares the reconstruction results of MaRO-GS and baseline 3DGS under each segmentation model. Under the same input masks, baseline 3DGS produces more noticeable background artifacts, whereas MaRO-GS suppresses these artifacts more effectively across all segmentation models. These results suggest that MaRO-GS is more robust than baseline 3DGS under different segmentation models.

\section{Seed Variance}
\label{supple_seed}

We repeat the key comparisons over three random seeds to assess run-to-run variation, and report the mean and standard deviation in Tables~\ref{tab:seed_variance_main} and~\ref{tab:seed_variance_dtu}.
For LERF-Mask and Mip-NeRF 360, we compare MaRO-GS with FDS-GS, which has the strongest competing PSNR in Table~1 of the main paper.
For Tanks and Temples, we report all methods because the PSNR margins among the compared methods are relatively small.
For DTU, we compare MaRO-GS with ObjectGS, which has the strongest competing PSNR across the Easy, Medium, and Hard splits in Table~3 of the main paper.

For MaRO-GS, the PSNR standard deviation ranges from 0.05 to 0.39 dB across the evaluated settings.
For each reported setting, the mean PSNR margin between MaRO-GS and the strongest competing method is larger than the PSNR standard deviation of either method.

\begin{table}[h]
\centering
\caption{Seed variance for selected methods from Table~1 in the main paper.
Mean $\pm$ standard deviation over three random seeds (180, 193, 5873).}
\label{tab:seed_variance_main}
\setlength{\tabcolsep}{6pt}
\resizebox{1.0\linewidth}{!}{%
\begin{tabular}{l|l|ccc}
\toprule
Dataset & Method & PSNR$\uparrow$ & SSIM$\uparrow$ & LPIPS$\downarrow$ \\
\midrule
\multirow{2}{*}{LERF-Mask} & FDS-GS & 40.76 $\pm$ 0.38 & 0.991 $\pm$ 0.0001 & 0.009 $\pm$ 0.0001 \\
 & Ours & 42.90 $\pm$ 0.14 & 0.992 $\pm$ 0.0000 & 0.008 $\pm$ 0.0000 \\
\midrule
\multirow{2}{*}{Mip-NeRF 360} & FDS-GS & 33.19 $\pm$ 0.29 & 0.977 $\pm$ 0.0012 & 0.024 $\pm$ 0.0008 \\
 & Ours & 34.64 $\pm$ 0.11 & 0.979 $\pm$ 0.0001 & 0.022 $\pm$ 0.0000 \\
\midrule
\multirow{5}{*}{Tanks \& Temples} & 3DGS & 30.68 $\pm$ 0.13 & 0.964 $\pm$ 0.0002 & 0.049 $\pm$ 0.0001 \\
 & PUP 3D-GS & 30.07 $\pm$ 0.08 & 0.955 $\pm$ 0.0002 & 0.070 $\pm$ 0.0001 \\
 & ObjectGS & 29.80 $\pm$ 0.44 & 0.957 $\pm$ 0.0002 & 0.054 $\pm$ 0.0007 \\
 & FDS-GS & 30.47 $\pm$ 0.15 & 0.966 $\pm$ 0.0004 & 0.042 $\pm$ 0.0004 \\
 & Ours & 31.17 $\pm$ 0.15 & 0.962 $\pm$ 0.0001 & 0.052 $\pm$ 0.0000 \\
\bottomrule
\end{tabular}
}
\end{table}

\begin{table}[h]
\centering
\caption{Seed variance for selected methods from Table~3 in the main paper.
Mean $\pm$ standard deviation over three random seeds (180, 193, 5873).}
\label{tab:seed_variance_dtu}
\setlength{\tabcolsep}{6pt}
\resizebox{1.0\linewidth}{!}{%
\begin{tabular}{l|l|ccc}
\toprule
Split & Method & PSNR$\uparrow$ & SSIM$\uparrow$ & LPIPS$\downarrow$ \\
\midrule
\multirow{2}{*}{Easy} & ObjectGS & 31.55 $\pm$ 0.11 & 0.967 $\pm$ 0.0008 & 0.056 $\pm$ 0.0004 \\
 & Ours & 33.26 $\pm$ 0.39 & 0.963 $\pm$ 0.0001 & 0.055 $\pm$ 0.0003 \\
\midrule
\multirow{2}{*}{Medium} & ObjectGS & 23.27 $\pm$ 0.59 & 0.903 $\pm$ 0.0038 & 0.116 $\pm$ 0.0043 \\
 & Ours & 26.50 $\pm$ 0.05 & 0.939 $\pm$ 0.0012 & 0.085 $\pm$ 0.0006 \\
\midrule
\multirow{2}{*}{Hard} & ObjectGS & 21.15 $\pm$ 0.43 & 0.913 $\pm$ 0.0034 & 0.106 $\pm$ 0.0018 \\
 & Ours & 22.95 $\pm$ 0.10 & 0.923 $\pm$ 0.0003 & 0.085 $\pm$ 0.0005 \\
\bottomrule
\end{tabular}
}
\end{table}

\section{More Qualitative Results}
\label{sec:H_0}

\begin{figure*}[!t]
    \centering
    \includegraphics[width=1.0\linewidth]{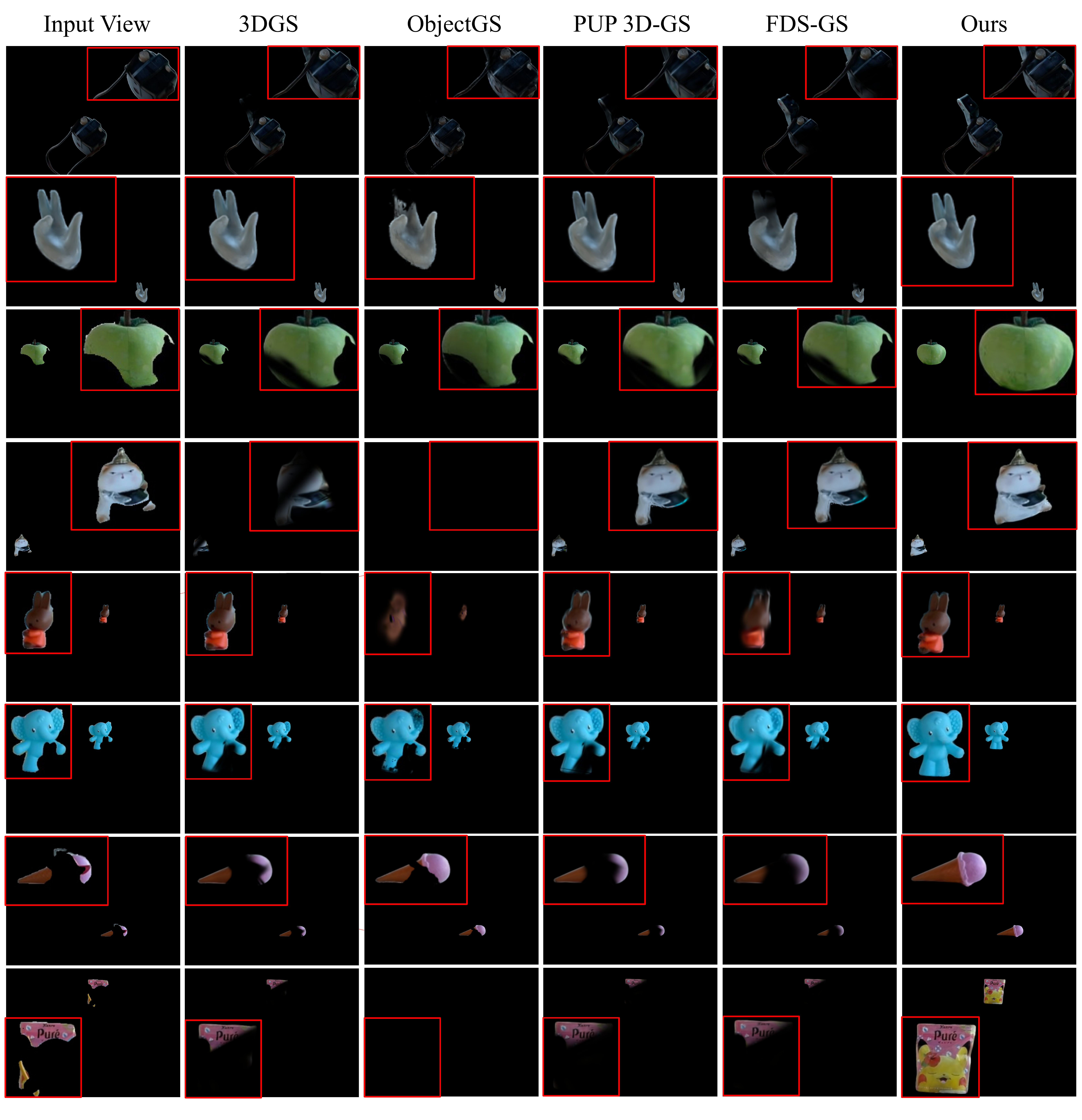} \\
    \vspace{-0.1cm} %
    \includegraphics[width=1.0\linewidth]{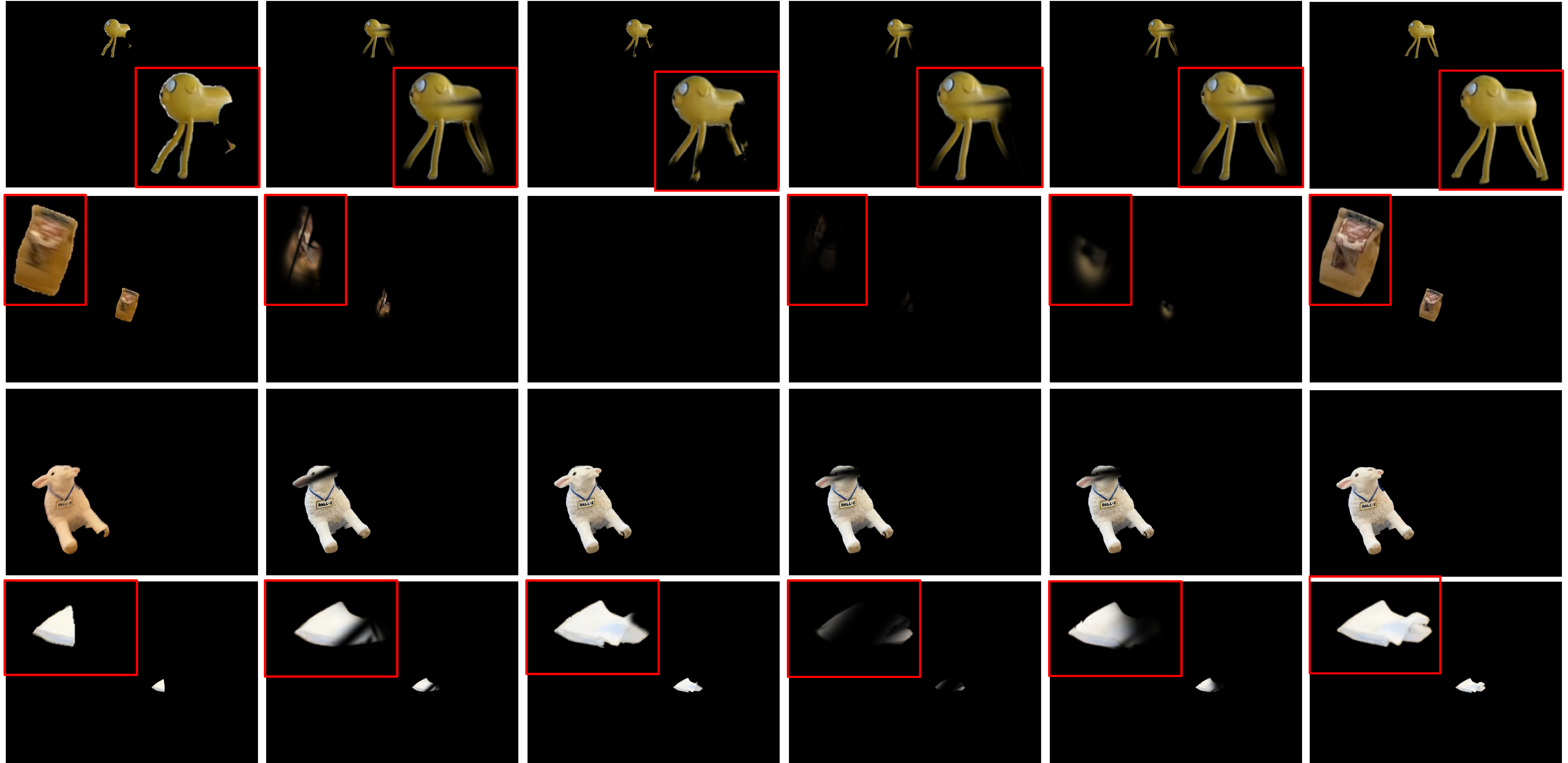}
    \caption{Visualization of object reconstruction on the LERF-Mask figurines scene. Red boxes denote the target object selected for reconstruction.}
    \label{fig_quality_lerf_1}
\end{figure*}

\begin{figure*}[!t]
    \centering
    \includegraphics[width=1.0\linewidth]{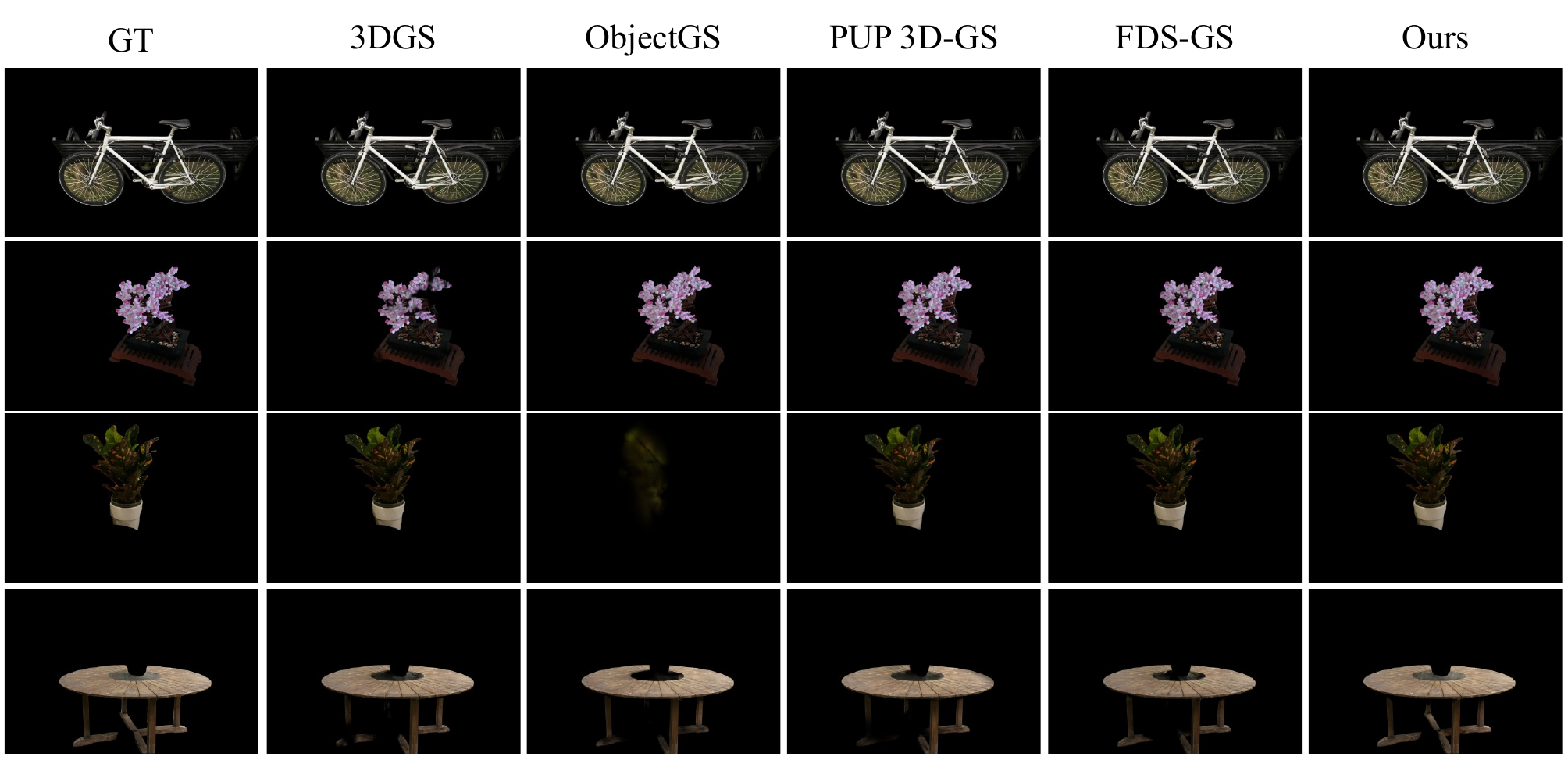}
    \caption{Visualization of object reconstruction on the Mip-NeRF scene.}
    \label{fig_quality_mip_nerf}
\end{figure*}

\begin{figure*}[!t]
    \centering
    \includegraphics[width=1.0\linewidth]{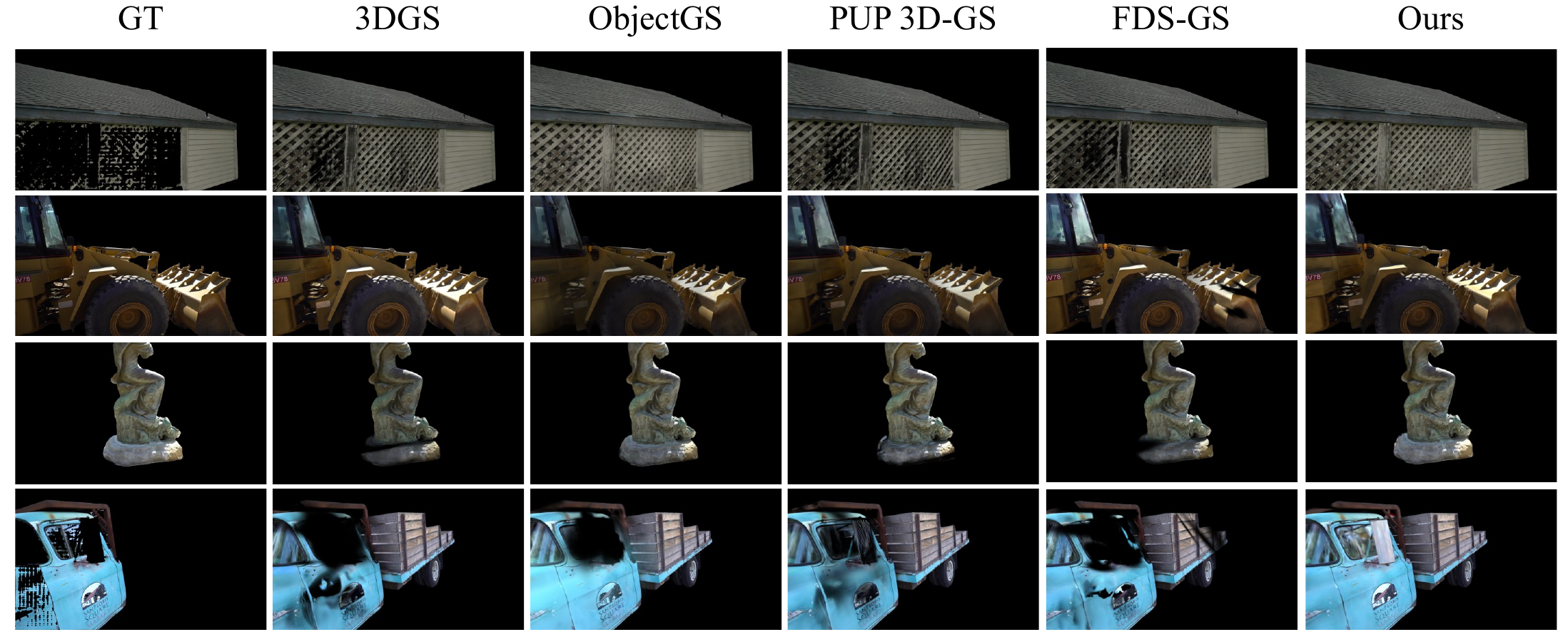}
    \caption{Visualization of object reconstruction on the Tanks and Temples scene.}
    \label{fig_quality_tanks} 
\end{figure*}

We provide additional qualitative reconstruction results across diverse scenes. \cref{fig_quality_lerf_1,fig_quality_mip_nerf,fig_quality_tanks} present object-centric reconstruction results on the LERF-Mask~\cite{Ye2024GaussianGrouping}, Mip-NeRF 360~\cite{mipnerf360}, and Tanks and Temples~\cite{knapitsch2017tanks} datasets, respectively. 

\end{document}